\documentclass[journal]{IEEEtran}
\usepackage{ifpdf}
\usepackage{cite}
\ifCLASSINFOpdf
   \usepackage[pdftex]{graphicx}
\else
\fi
\usepackage{amsmath}
\usepackage{algorithmic}
\usepackage{array}
\usepackage{url}
\usepackage{booktabs} % for professional tables
\usepackage{amsfonts,amssymb}
\usepackage{subfigure}
\usepackage{color,xcolor}
\usepackage{times}
\usepackage{epsfig}
\usepackage{amsmath}
\usepackage{textcomp}
\usepackage{gensymb}
\usepackage{wasysym}
\usepackage{multirow}
\usepackage{cite}
\usepackage{graphicx} % more modern

\usepackage{algorithm}
\usepackage{algorithmic}
\usepackage{mathrsfs}
\usepackage{url}
\usepackage{lettrine}

\usepackage{algorithm}
\usepackage{subfigure}
\newtheorem{theorem}{Theorem}

\usepackage{verbatim}
\usepackage{bm}

\begin{document}
%
% paper title
% Titles are generally capitalized except for words such as a, an, and, as,
% at, but, by, for, in, nor, of, on, or, the, to and up, which are usually
% not capitalized unless they are the first or last word of the title.
% Linebreaks \\ can be used within to get better formatting as desired.
% Do not put math or special symbols in the title.
\title{Training Robust Deep Neural Networks via Adversarial Noise Propagation}
%
%
% author names and IEEE memberships
% note positions of commas and nonbreaking spaces ( ~ ) LaTeX will not break
% a structure at a ~ so this keeps an author's name from being broken across
% two lines.
% use \thanks{} to gain access to the first footnote area
% a separate \thanks must be used for each paragraph as LaTeX2e's \thanks
% was not built to handle multiple paragraphs
%
%
%\IEEEcompsocitemizethanks is a special \thanks that produces the bulleted
% lists the Computer Society journals use for "first footnote" author
% affiliations. Use \IEEEcompsocthanksitem which works much like \item
% for each affiliation group. When not in compsoc mode,
% \IEEEcompsocitemizethanks becomes like \thanks and
% \IEEEcompsocthanksitem becomes a line break with idention. This
% facilitates dual compilation, although admittedly the differences in the
% desired content of \author between the different types of papers makes a
% one-size-fits-all approach a daunting prospect. For instance, compsoc
% journal papers have the author affiliations above the "Manuscript
% received ..."  text while in non-compsoc journals this is reversed. Sigh.
\author{
Aishan Liu, Xianglong~Liu*, Chongzhi Zhang$^\dag$, Hang Yu$^\dag$, Qiang Liu, Dacheng Tao

\thanks{A. Liu, X. Liu, C. Zhang, and H. Yu are with the State Key Lab of Software Development Environment, Beihang University, Beijing 100191, China. X. Liu is also with Beijing Advanced Innovation Center for Big Data-Based Precision Medicine (*Corresponding author: Xianglong Liu, xlliu@nlsde.buaa.edu.cn), ($\dag$ indicates equal contributions, authors in alphabetical order)}

\thanks{Q. Liu is with the Department of Computer Science, University of Texas at Austin, Austin, TX 78712, USA}

%\thanks{Q. Wu is with the College of Information Engineering, Henan University of Science and Technology, Luoyang 471000, China}

\thanks{D. Tao is with the UBTECH Sydney Artificial Intelligence Centre and the School of Information Technologies, Faculty of Engineering and Information Technologies, The University of Sydney, Darlington, NSW 2008, Australia}
}
% <-this % stops a space

% The paper headers
\markboth{IEEE Transactions on Image Processing,~Vol.~, No.~, December~2020}%
{Shell \MakeLowercase{\textit{Liu et al.}}: Training Robust Deep Neural Networks via Adversarial Noise Propagation}
% The only time the second header will appear is for the odd numbered pages
% after the title page when using the twoside option.
%
% *** Note that you probably will NOT want to include the author's ***
% *** name in the headers of peer review papers.                   ***
% You can use \ifCLASSOPTIONpeerreview for conditional compilation here if
% you desire.

% The publisher's ID mark at the bottom of the page is less important with
% Computer Society journal papers as those publications place the marks
% outside of the main text columns and, therefore, unlike regular IEEE
% journals, the available text space is not reduced by their presence.
% If you want to put a publisher's ID mark on the page you can do it like
% this:
%\IEEEpubid{0000--0000/00\$00.00~\copyright~2015 IEEE}
% or like this to get the Computer Society new two part style.
%\IEEEpubid{\makebox[\columnwidth]{\hfill 0000--0000/00/\$00.00~\copyright~2015 IEEE}%
%\hspace{\columnsep}\makebox[\columnwidth]{Published by the IEEE Computer Society\hfill}}
% Remember, if you use this you must call \IEEEpubidadjcol in the second
% column for its text to clear the IEEEpubid mark (Computer Society jorunal
% papers don't need this extra clearance.)

% use for special paper notices
%\IEEEspecialpapernotice{(Invited Paper)}

% for Computer Society papers, we must declare the abstract and index terms
% PRIOR to the title within the \IEEEtitleabstractindextext IEEEtran
% command as these need to go into the title area created by \maketitle.
% As a general rule, do not put math, special symbols or citations
% in the abstract or keywords.
\IEEEtitleabstractindextext{%
\begin{abstract}
In practice, deep neural networks have been found to be vulnerable to \textcolor{black}{various types of noise}, such as adversarial examples and corruption. Various adversarial defense methods have accordingly been developed to improve adversarial robustness for deep models. However, simply training on data mixed with adversarial examples, most of these models still fail to defend against the generalized types of \textcolor{black}{noise}. Motivated by the fact that hidden layers play a highly important role in maintaining a robust model, this paper proposes a simple yet powerful training algorithm, named \emph{Adversarial Noise Propagation} (ANP), which injects \textcolor{black}{noise} into the hidden layers in a layer-wise manner. ANP can be implemented efficiently by exploiting the nature of the backward-forward training style. \textcolor{black}{Through thorough investigations, we determine that different hidden layers make different contributions to model robustness and clean accuracy, while shallow layers are comparatively more critical than deep layers. Moreover, our framework can be easily combined with other adversarial training methods to further improve model robustness by exploiting the potential of hidden layers.} Extensive experiments on MNIST, CIFAR-10, CIFAR-10-C, CIFAR-10-P, and ImageNet demonstrate that ANP enables the strong robustness for deep models against both adversarial and corrupted ones, and also significantly outperforms various adversarial defense methods.
\end{abstract}

% Note that keywords are not normally used for peerreview papers.
}

% make the title area
\maketitle

% To allow for easy dual compilation without having to reenter the
% abstract/keywords data, the \IEEEtitleabstractindextext text will
% not be used in maketitle, but will appear (i.e., to be "transported")
% here as \IEEEdisplaynontitleabstractindextext when the compsoc
% or transmag modes are not selected <OR> if conference mode is selected
% - because all conference papers position the abstract like regular
% papers do.
\IEEEdisplaynontitleabstractindextext
% \IEEEdisplaynontitleabstractindextext has no effect when using
% compsoc or transmag under a non-conference mode.

% For peer review papers, you can put extra information on the cover
% page as needed:
% \ifCLASSOPTIONpeerreview
% \begin{center} \bfseries EDICS Category: 3-BBND \end{center}
% \fi
%
% For peerreview papers, this IEEEtran command inserts a page break and
% creates the second title. It will be ignored for other modes.
\IEEEpeerreviewmaketitle

\section{Introduction}\label{sec:introduction}
% Computer Society journal (but not conference!) papers do something unusual
% with the very first section heading (almost always called "Introduction").
% They place it ABOVE the main text! IEEEtran.cls does not automatically do
% this for you, but you can achieve this effect with the provided
% \IEEEraisesectionheading{} command. Note the need to keep any \label that
% is to refer to the section immediately after \section in the above as
% \IEEEraisesectionheading puts \section within a raised box.
\IEEEPARstart{R}{ecent} advances in deep learning have achieved remarkable successes in various challenging tasks, including computer vision \cite{Krizhevsky2012ImageNet,7485869,7839189}, natural language processing \cite{bahdanau2014neural,rajadesingan2015sarcasm} and speech \cite{Hinton2012Deep,amodei2016deep}. In practice, deep learning has been routinely applied on large-scale datasets containing data collected from daily life, which inevitably contain large amounts of \textcolor{black}{noise} including adversarial examples and corruption \cite{szegedy2013intriguing,goodfellow6572explaining}. Unfortunately, while such \textcolor{black}{noise} is imperceptible to human beings, it is highly misleading to deep neural networks, which presents potential security threats for practical machine learning applications in both the digital and physical world \cite{papernot2016practical,Liu2019Perceptual,8423654, Liu2020Spatiotemporal, Liu2020Biasbased}.

%Recently,
%Adversarial robustness has become one of the most active research topics.
Over the past few years, \textcolor{black}{the training of robust deep neural networks} against \textcolor{black}{noise} has attracted significant attention. The most successful strategy has tended to involve developing different adversarial defense strategies \cite{xie2018mitigating,dhillon2018stochastic,buckman2018thermometer,guo2017countering,song2017pixeldefend,8417973,Zhang2020Interpreting} against adversarial examples. A large proportion of these defensive methodologies attempt to supply adversaries with non-computable gradients to avoid common gradient-based adversarial attacks. \textcolor{black}{While they can obtain a certain degree of stabilization for DNNs in adversarial setting, these methods can be easily circumvented by constructing a function to approximate the non-differentiable
layer on the backward pass. \cite{athalye2018obfuscated}}. By contrast, \emph{adversarial training} \cite{szegedy2013intriguing} can still mount a appropriate defense by augmenting training data with adversarial examples. However, while adversarially trained deep models \cite{alexey2017adversarialmachine} are robust to some single-step attacks, they remain vulnerable to iterative attacks. More recently, \cite{NIPS2018_7324} proposed to improve adversarial robustness by integrating an adversarial perturbation-based regularizer into the classification objective.
 %a perturbation-based regularizer which tries to make the model generate adversarial examples harder by enlarging $\Delta x$.
 %integrate an adversarial perturbation-based regularizer into the classification objective, such that the obtained models learn to resist potential attacks.
 %directly and precisely. The whole optimization problem is solved just like training a recursive network.

In addition to adversarial examples \cite{xie2018mitigating,pmlr_v70_cisse17a}, corruption such as snow and blur also frequently occur in the real world, which also presents critical challenges for the building of strong deep learning models. \cite{dodge2017study} found that deep learning models behave distinctly subhuman to input images with Gaussian \textcolor{black}{noise}. \cite{zheng2016improving} proposed stability training to improve model robustness against \textcolor{black}{noise}, but this was confined only to JPEG compression. More recently, \cite{hendrycks2018benchmarking} was the first to establish a rigorous benchmark to evaluate the robustness of image classifier to 75 different types of corruption.%techniques are used to create test images.
%Therefore, it is extraordinarily significant to study more about adversarial examples as well as corruption, thereby create deep learning models that are robust to both of them, especially when models are deployed in safety-critical scenarios.

Despite the progress already achieved, few studies have been devoted to improving model robustness against corruption\textcolor{black}{. Most} existing adversarial defense methods remain vulnerable to the generalized \textcolor{black}{noise}, which is mainly due to the use of a simple training paradigm that adds adversarial \textcolor{black}{noise} to the input data. It is well understood that, in deep neural networks, the influence of invasive \textcolor{black}{noise} on prediction can be observed directly in the form of sharp variations in the middle feature maps during forward propagation \cite{liao2018defense,pmlr_v70_cisse17a}. Prior studies \cite{pmlr_v70_cisse17a} have proven that hidden layers play a very important role in maintaining a robust model. In \cite{pmlr_v70_cisse17a}, adversarially robust models can be created by constraining the Lipschitz constant between hidden layers (e.g., linear, conv layers) to be smaller than 1. \cite{ilyas2019adversarial} generates robust features with the help of the penultimate layer of a classifier, which are in turn used to help with training a robust model. More recently, \cite{santurkar2019computer} also noted the importance of robust feature representation and used this approach to deal with many computer vision tasks. This indicates that the noise resistance of hidden layers plays a highly important role in training robust models. \textcolor{black}{Motivated by this fact, we aim to build strong deep models by obtaining robust hidden representations during training.}

Accordingly, in this paper, we propose a simple but very powerful training algorithm named \emph{Adversarial Noise Propagation} (ANP), designed to enable strong robustness for deep models against generalized \textcolor{black}{noise} (including both adversarial examples and corruption). Rather than perturbing the inputs, as in traditional adversarial defense methods, our method injects adversarial \textcolor{black}{noise} into the \emph{hidden layers} of neural networks during training. This can be accomplished efficiently by a simple modification of the standard backward-forward propagation, without introducing significant computational overhead. \textcolor{black}{Since the adversarial perturbations for the hidden layers are considered and added during training, models trained using ANP are expected to be more robust against more types of noise.} Moreover, ANP takes advantage of hidden layers and is orthogonal to most adversarial training methods; \textcolor{black}{thus, they could be combined together to build stronger models.} To facilitate further understanding of the contributions of hidden layers, we provide insights into their behaviors during training from the perspectives of hidden representation insensitivity and human visual perception alignment. Our code has also been released at \url{https://github.com/AnonymousCodeRepo/ANP}.

\textcolor{black}{Extensive experiments in both black-box and white-box settings on MNIST, CIFAR-10 and ImageNet are conducted to demonstrate that ANP is able to achieve excellent results compared to the common adversarial defense algorithms, including the adversarial defense methods won at \emph{NeurIPS 2017}. Meanwhile, experiments on CIFAR-10-C and CIFAR-10-P \cite{hendrycks2018benchmarking} prove that ANP can enhance strong corruption robustness for deep models. By investigating the contributions of hidden layers, we found that we only need to inject noise into shallow layers, which are more critical to model robustness. In addition, we can further improve model robustness by combining ANP with other adversarial training methods in different settings.}

% The very first letter is a 2 line initial drop letter followed
% by the rest of the first word in caps (small caps for compsoc).
%
% form to use if the first word consists of a single letter:
% \IEEEPARstart{A}{demo} file is ....
%
% form to use if you need the single drop letter followed by
% normal text (unknown if ever used by the IEEE):
% \IEEEPARstart{A}{}demo file is ....
%
% Some journals put the first two words in caps:
% \IEEEPARstart{T}{his demo} file is ....
%
% Here we have the typical use of a "T" for an initial drop letter
% and "HIS" in caps to complete the first word.
%\IEEEPARstart{T}{his} demo file is intended to serve as a ``starter file''
%for IEEE Computer Society journal papers produced under \LaTeX\ using
%IEEEtran.cls version 1.8b and later.
% You must have at least 2 lines in the paragraph with the drop letter
% (should never be an issue)
%I wish you the best of success.

%\hfill mds

%\hfill August 26, 2015
{\color{black}{
\section{Related Works}
A number of works have been proposed to improve adversarial robustness, including those utilizing network distillation \cite{Papernot2015Distillation}, input reconstruction \cite{Gu2014Towards,song2018pixeldefend}, gradient masking \cite{xie2018mitigating,Wang2016Learning}, etc. Among these, adversarial training \cite{goodfellow6572explaining,madry2017towards,kurakin2016adversarial} in particular has been widely studied in the adversarial learning literature and determined to be the most effective method for improving model robustness against adversarial examples.
The concept of adversarial training, which was first proposed by \cite{goodfellow6572explaining}, involves feeding model with adversarial examples to facilitate data augmentation during training:

\begin{align*}
\min_{\theta}\rho(\theta), \quad \rho(\theta)= \mathbb{E}_{(x,y)\sim D}\left[\max_{r \in S}L(y, F(x+r , \theta)) \right],
\end{align*}

\noindent where $r$ is a small ball that controls the magnitude of the noise.

These authors also proposed a gradient-based attack method, called FGSM, to generate adversarial examples for adversarial training. To further improve the effectiveness of adversarial training, PGD-based adversarial training \cite{madry2017towards} was subsequently introduced to adversarially train deep models via PGD attack \cite{madry2017towards}. It can be readily seen that both FGSM- and PGD-based adversarial training only consider adversarial \textcolor{black}{noise} in the input data; thus they can be considered a special case of ANP in which only adversarial noise in the $0$-th hidden layer is considered.
%Furthermore, the standard PAT and NAT typically only perform one step of gradient ascent on the adversarial noise, corresponding to the case of $k=1$ in our framework.

To improve the diversity of adversarial examples, ensemble adversarial training \cite{tramer2017ensemble} was devised; this approach employs a set of models $F$ to generate different adversarial examples for data augmentation. In ANP, various types of adversarial \textcolor{black}{noise} (with different noise sizes, iteration steps, and noise magnitudes) are generated in each layer, which could increase the diversity and complexity of adversarial \textcolor{black}{noise} injected into the model.

Layer-wise adversarial training \cite{sankaranarayanan2018regularizing} was proposed as a regularization mechanism to prevent overfitting. During training, adversarial gradients in the current mini-batch are computed with reference to the previous mini-batch. Their method is primarily designed to improve the model's generalization ability, which is a different goal from ours. For ANP, the adversarial \textcolor{black}{noise} for one specified mini-batch is computed only in the same mini-batch. We believe that the high correlation of adversarial gradients within a single mini-batch will lead to better performance. We therefore introduce the progressive backward-forward propagation in order to fully utilize the information contained in every group of mini-batch data. We further prove that only shallow layers should be considered during training to obtain a robust model.

In summary, our ANP can be considered as a more general framework that considers the contribution of hidden layers and observes adversarial \textcolor{black}{noise} in more flexible ways.}}

\section{Preliminaries}

\subsection{Terminology and Notation}
%Assume $\mathcal{X}$ $\subseteq  \mathbb{R}^{n}$ is the feature space with $n$ the number of features. Suppose
Given a dataset with feature vector $x$ $\in$ $\mathcal{X}$ and
label $y$ $\in$ $\mathcal{Y}$, % the corresponding class label.
the deep supervised learning model aims to learn a mapping or classification function $f$: $\mathcal{X}$ $\rightarrow$ $\mathcal{Y}$. More specifically, in this paper, we consider the visual recognition problem.

\subsection{Adversarial Example}
Given a network $f_{\theta}$ and an input $x$ with ground truth label $y$, an adversarial example $x^{adv}$ is an input such that
\begin{align*}
f_{\theta}(x^{adv}) \neq y  \quad s.t. \quad \|x-x^{adv}\| < \epsilon,
\end{align*}
where $\|\cdot\|$ is a distance metric used to quantify that the semantic distance between the two inputs $x$ and $x^{adv}$ is small enough. By contrast, the adversarial example makes the model predict the wrong label: namely, $f_{\theta}(x^{adv}) \neq y$.

\subsection{Corruption}
Image corruption refers to random variations of the brightness or color information in images, such as Gaussian noise, defocus blur, brightness, etc.
Supposing, we have a set of corruption functions $C$ in which each $c(x)$ performs a different kind of corruption function. Thus, average-case model performance on small, general, classifier-agnostic corruption can be used to define model corruption robustness as follows
\begin{align*}
\mathbb{E}_{c \thicksim C}[P_{(x,y) \thicksim D}(f(c(x))=y)].
\end{align*}
%Average-case model performance on small, general, classifier-agnostic perturbations define model perturbation robustness:
%\begin{align*}
%\mathbb{E}_{\varepsilon \thicksim E}[P_{(x,y) \thicksim D}(f(\epsilon(x)=y))].
%\end{align*}
In summary, corruption robustness measures the classifier's average-case performance on corruption $C$, while adversarial robustness measures the worst-case performance on small, additive, classifier-tailored perturbations.

\section{Proposed Approach}
In this section, we introduce our proposed approach, \emph{Adversarial Noise Propagation} (ANP).% The optimization details and the training process will be presented.
%\subsection{Corruption and Perturbation}
\subsection{Adversarial Formulation}
%\textcolor{black}{(***introduce the concept and add more content about the perturbation propagation in this section, ***)}
\textcolor{black}{In a deep neural network, the sharp variations in the hidden representation will propagate through hidden layers, leading to undesired predictions.} Therefore, model robustness can be greatly improved by the noise insensitivity and guaranteeing the stable behavior in hidden layers. To obtain a model that is robust to small degress of noise, \textcolor{black}{we try to improve the layer-wise noise resistance ability in the deep learning models.}

\textcolor{black}{Instead of manipulating only the input layer, as in traditional adversarial defense methods, we instead attempt to add adversarial \textcolor{black}{noise} to each hidden layer of a deep learning model by propagating backward from the adversarial loss during training.} This strategy forces the model to minimize the model loss for a specific task, exploiting the opposite adversarial noise in each hidden layer that expects to maximize the loss. Subsequently, the learned parameters in each layer enable the model to maintain consistent and stable predictions for the clean instance and its noisy surrogates distributed in the neighborhood, thus building strong robustness for deep models.

From a formal perspective, let us first recall that a deep neural network $ y = F(x; \theta)$
is a composition of a number of nonlinear maps, each of which corresponds to a layer:
\begin{align*}
z_{m+1} = f(z_{m}; ~\theta),\ m = 0,\ldots, M,
\end{align*}
where $z_0 = x$ denotes the input, $z_{M} = y$ the output, and $z_m$ the output of the $m$-th hidden layer. Moreover, $\theta$ collects the weights of the network.

In our framework,
we introduce an adversarial noise $r_m$
on the hidden state $z_{m}$ at each layer, as follows:
\begin{align*}
z_{m+1} = f(z_m + r_m, ~\theta),\ m = 0,\ldots, M.
\end{align*}
We use $\tilde y = F(x;~ \theta, r)$ to denote the final network output with the injected \textcolor{black}{noise} $r = \{r_m\}_{m=0}^M$ at all layers.
%
%Therefore, to learn a robust deep models, our adversarial training framework should satisfy the following formulation
We then learn the network parameter $\theta$ by minimizing the following adversarial loss:
{\color{black}{
\begin{equation}
\begin{aligned} \label{overall_eq}
%\min_\theta &\mathbb{E}_{(x,y)\thicksim D} \left [  \max_{r } \left ( L(y, ~ F(x; ~ \theta, r)) - \frac{\eta}{2} ||r||_2^2  \right ) \right ],
%\\& s.t. \quad R = \{ r \colon ||r^m||_2 < \varepsilon, \quad 0 \leq m \leq M\},
\min_\theta &\mathbb{E}_{(x,y)\thicksim D} \left [  \max_{r } \left ( L(y, ~ F(x; ~ \theta, r)) - \eta\cdot ||r||_p  \right ) \right ],
\end{aligned}
\end{equation}
}}
where, for each data point $(x,y)$, we search for an adversarial noise $r$, subject to an \textcolor{black}{$\ell_p$} norm constraint. The coefficient $\eta$ controls the magnitude of the adversarial noise.
%where $R$ denotes the feasible region for the adversarial noise, and we empirically take a $L_2$ ball. %Note that we can hardly solving the above problem directly the adversarial noise is optimized for each data point individually.

%Note that each data poin%t $(x,y)$ is associated with its
%\subsection{Learning with the noise register}
\subsection{Noise Propagation}
Evidently, the key challenge of this framework is solving the inner maximization for individual input data points. This is efficiently addressed in our \emph{Adversarial Noise Propagation} (ANP)
%with %developed to efficiently pursue the solution in
method by performing gradient descent on $r$, utilizing a natural backward-forward style training that adds minimum computational cost over the standard back-propagation. This induces \textcolor{black}{noise} propagation across layers and noise injection into hidden layers during the backward-forward training.
%$%that perform simple

More specifically, in each iteration, we first select a mini-batch of training data. For each data point $(x,y)$ in the mini-batch, we approximate the inner optimization by running $k$ steps of gradient descent to utilize the most information possible in each mini-batch. After initializing at $r^{m,0} = 0$ for all $m=0,\ldots,M$, we have the adversarial gradient for the $m$-th hidden layer $z^m$:
\begin{equation}\label{eq:gd}
  g^{m,t}
  = \nabla_{r^{m}} L(y, F(x; ~\theta, r^{t})) =  \nabla_{z^{m}} L(y, F(x; ~\theta, r^{t})),\nonumber
 \end{equation}
from which we can make the key observation that the gradient of the adversarial noise $r^m$ is equal to the gradient of the hidden states $z^m$. \textcolor{black}{Evidently, this has already been calculated in the standard backward propagation, and thus introduces no additionally computational cost.}

More specifically, the noise gradient is calculated recursively during the standard backward propagation, as follows: %standard
 %However, when it comes to the calculation of the adversarial gradient for the $m$-th hidden layer, Equation (\ref{eq:gd}) cannot be utilized directly due to the expensive computation. But fortunately, we have the chain-rule to simplify the computation as follows:
\begin{align*}
%=\nabla_{z^{m-1}}J(y|x,\theta)
g^{m-1,t}= &\frac{\partial L(y, F(x; \theta, r)}{\partial z^{m-1,t}}\\
&=\frac{\partial L(y, F(x; \theta, r)}{\partial z^{m,t}} \cdot \frac{\partial z^{m,t}}{\partial z^{m-1,t}}\\
&=g^{m,t} \cdot \frac{\partial z^{m,t}}{\partial z^{m-1,t}}.
\end{align*}

Performing gradient descent on $r^m$ yields a further update, namely:
%In order to control the feasible magnitude for adversarial noises, we further adopt the noise regularization coefficient $\eta$. Furthermore, $\eta$ also controls the amount of noises the previous step contributes to the current step. By introducing step size $\varepsilon$, the noises added in each step are confined and the $L_2$ ball constraint of overall noise is satisfied. Thus, the adversarial noise can be propagated to $m$-th hidden layer following the adversarial gradient:
{ \color{black}{
 \begin{equation}
\begin{aligned}% \label{eq}
& r^{m,t+1} \gets (1-\eta) r^{m,t} +  \frac{\varepsilon}{k} \frac{g^{m,t}}{||g^{m,t}||_p},
\end{aligned}
 \end{equation}
 }}
 where $\varepsilon$ is the step size; moreover, it is normalized by the number of steps $k$,
 so that the overall magnitude of update for $k$ steps equals $\varepsilon$. Practically speaking, $\eta$ and $\varepsilon$ control the contribution made by the previous noise value $r^{m,t}$ and the gradient $g^{m,t}$ to the new noise $r^{m,t+1}$, respectively.
 %, which controls the noise magnitude of $r.$
%Also, $r^{m,t+1}$ accumulates adversarial gradients from $r^{m,t}$ which help to overcome local optima.

%However, when it comes to the calculation of the adversarial gradient for the $m$-th hidden layer, Equation (\ref{eq:gd}) cannot be utilized directly due to the expensive computation. But fortunately, we have the chain-rule to simplify the computation as follows:
%\begin{align*}
%=\nabla_{z^{m-1}}J(y|x,\theta)
%g^{m-1}&=\frac{\partial J(y|x,\theta)}{\partial z^{m-1}}=\frac{\partial J(y|x,\theta)}{\partial z^{m}} \cdot \frac{\partial z^m}{\partial z^{m-1}}\\
%%&=g^m \cdot \frac{\partial z^m}{\partial z^{m-1}}.
%\end{align*}

\subsubsection{Learning with a Noise Register}
\begin{figure}[t!]
	\centering
	%\vspace{-0.3in}
	\includegraphics[width=1.1\linewidth]{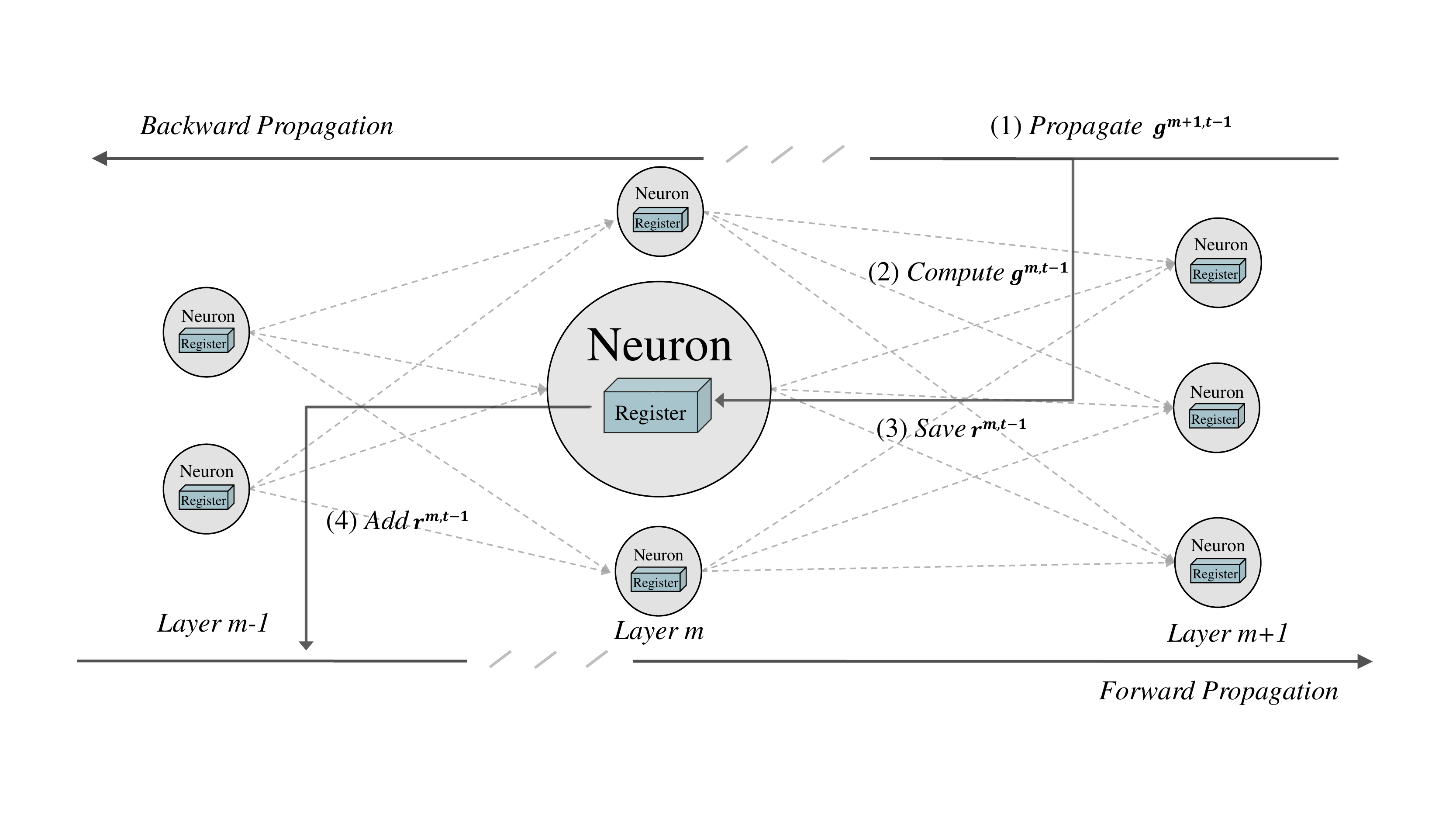}
	%\vspace{-.4in}
	\caption{Adversarial noise propagation with the noise register during backward-forward training.}
	\label{fig:fig1}
%\vspace{-.15in}
\end{figure}

In practice, ANP is implemented efficiently through a backward-forward training process. In more detail, for each mini-batch training, we store the corresponding adversarial noise $r$ for each hidden layer during the backward propagation process; during the forward propagation process, moreover, we simply fetch $r$ as the noise and add it to the input in the corresponding hidden layer following the affine transformation and before the activation function. This training procedure introduces no substantial increase in either computation and memory consumption, except that we need to integrate a register $S$ into each neuron in order to store the adversarial noise $r$ (as illustrated in Figure \ref{fig:fig1}). At the time of inference, $S$ can be discarded, meaning that it has no influence on the model complexity. \textcolor{black}{For example, for the $m$-th hidden layer, during the forward propagation process at the $t$-th iteration, we first compute the affine transformation $z^{m,t}$=$a^{m-1,t}  w^{m-1}$ + $b^{m-1} $. More specifically, $w^{m-1}$ and $b^{m-1}$ denote the weight and bias for the affine transformation, and $a^{m-1,t}$ represents the activation at the previous layer. We next fetch and add the adversarial \textcolor{black}{noise} $r^{m,t}$ to $z^{m,t}$, and compute the activation $a^{m,t}=relu(z^{m,t})$. The adversarial \textcolor{black}{noise} is subsequently propagated to the next hidden layer. In contrast to the traditional adversarial training methods, we feed only clean examples to the models during training. Adversarial noise for each layer is computed and propagated to train the robust models.} Algorithm \ref{alg:mini-batch} outlines more details of the training process with ANP in each mini-batch.

%\subsection{Neuron Structure}
%For the sake of reducing computation and time complexity, we use equation (11) to calculate adversarial gradients for each hidden layer.

%\subsubsection{Optimization with Progressive and Momentum}
%In order to utilize the most information in each mini-batch as well as fit data distribution better, progressive training process with momentum is introduced. Thus, $k$ times of training is needed for each mini-batch. Let $t$ denote the current training step for a specific mini-batch, and $\alpha$ controls the momentum decay factor:
%\begin{equation}
%\begin{aligned}
%r^{m,t}=\alpha \cdot r^{m,t-1} + \frac{g^{m,t-1}}{||g^{m,t-1}||_2}.
%\end{aligned}
%\end{equation}
%In forward propagation process, we add $r^{m,t}$ to corresponding hidden layer after affine transformation or convolution filter:
%\begin{equation}
%\begin{aligned}
%z^{m,t}=z^{m,t}+\beta \cdot r^{m,t},
%\end{aligned}
%\end{equation}
%where $\beta=\frac{\varepsilon}{k}$. For the sake of constraint on the perturbations added, equation () is modified as:
%\begin{equation}
%\begin{aligned}
%z^{m,t}=z^{m,t}+\beta \cdot \frac{r^{m,t}}{||r^{m,t}||_2},
%\end{aligned}
%\end{equation}
%so that overall noises injected satisfy $\Sigma_{t=1}^{k}\beta \cdot \frac{r^{m,t}}{||r^{m,t}||_2} < \varepsilon$.
%
%To sum up, the objective function for ANP is shown as follows:
%\begin{equation}
%\begin{aligned}
%\mathbb{L}_{ANP} &= \min_{\theta} \mathbb{E}_{(x,y)\in D}[J(y_{true}|z^{m,t}+\beta \cdot \frac{r^{m,t}}{||r^{m,t}||_2}, \theta)] \\&
%s.t. \quad r^{m,t}=\alpha \cdot r^{m,t-1} + \frac{g^{m,t-1}}{||g^{m,t-1}||_2} ,
%\\& \quad 0 \leq m \leq n ,\quad 1 \leq t \leq k .
%\end{aligned}
%\end{equation}

\begin{algorithm}[thb]
   \caption{Adversarial Noise Propagation (ANP)}
   \label{alg:mini-batch}
\hspace*{0.02in} {\bf Input:} mini-batch data $(x,y)$\\
\hspace*{0.02in} {\bf Output:} robust model parameters $\theta$\\
\hspace*{0.02in} {\bf Hyper-parameter:} $\eta, \varepsilon$ and $k$
\begin{algorithmic}[1]
   \FOR{$t$ in $k$ steps}
   \STATE // \textit{Backward propagation} \nonumber
    \STATE Update model parameters $\theta$ using
    standard back-propagation.
    %normal gradient descent
   %\STATE calculate and save adversarial noise for each hidden layer:
   \STATE Compute and propagate adversarial gradient:\\ $g^{m,t-1} = g^{m+1,t-1} \cdot \frac{\partial z^{m+1,t-1}}{\partial z^{m,t-1}}$
   \STATE Compute, propagate and save adversarial noise:\\ $r^{m,t} = (1-\eta) r^{m,t-1} + \frac{\varepsilon}{k} \frac{g^{m,t-1}}{||g^{m,t-1}||_p}$
   \STATE // \textit{Forward propagation} \nonumber
   % restore adversarial noise calculated at the last step for each hidden layer:
   \STATE Compute the affine transformation:\\ $z^{m,t}=a^{m-1,t} w^{m-1} + b^{m-1} $
   \STATE Fetch and add adversarial noise:\\  $z^{m,t} += r^{m,t}$
   \STATE Compute the activation:\\$a^{m,t}=relu(z^{m,t})$
   \ENDFOR
\end{algorithmic}
\end{algorithm}

\section{Experiments and Evaluation}
In this section, we will evaluate our proposed ANP on the popular image classification task. Following the guidelines from \cite{carlini2019evaluating}, we compare ANP with several state-of-the-art adversarial defense methods against both adversarial \textcolor{black}{noise} and corruption as well.

\subsection{Experimental Setup}

\textbf{Datasets and models}. To assess the adversarial robustness, we adopt the widely used {MNIST}, {CIFAR-10} and {ImageNet} datasets. MNIST is a dataset of 10 classes of handwritten digits of size $28 \times 28$, containing 60K training examples and 10K test instances \cite{lecun1998mnist}. We use LeNet for MNIST. CIFAR-10 consists of 60K natural scene color images, with 10 classes, of size $32 \times 32 \times 3 $ \cite{krizhevsky2009learning}. We further use VGG-16, ResNet-18, DenseNet and InceptionV2 for CIFAR-10. ImageNet contains 14M images with more than 20k classes \cite{deng2009imagenet}. In the interests of simplicity, we only choose 200 classes from the 1000 available in ILSVRC-2012, with 100K and 10k images used for training set and test set, respectively. The models we use for ImageNet are ResNet-18 and AlexNet.

\textbf{Adversarial attacks}. We apply a diverse set of adversarial attack algorithms including FGSM \cite{goodfellow6572explaining}, BIM \cite{kurakin2016adversarial}, Step-LL \cite{kurakin2016adversarial}, MI-FGSM \cite{dong2017boosting}, PGD \cite{madry2017towards}, and BPDA \cite{athalye2018obfuscated} in terms of $\ell_{\infty}$-norm. We also use C\&W \cite{carlini2017towards} in terms of $\ell_2$-norm.

\textbf{Corruption attacks}. To assess the corruption robustness, we test our proposed method on {CIFAR-10-C} and {CIFAR-10-P} \cite{hendrycks2018benchmarking}. These two datasets are the first choice for benchmarking model static and dynamic model robustness against different common corruption and noise sequences at different levels of severity \cite{hendrycks2018benchmarking}. They are created from the test set of CIFAR-10 using 75 different corruption techniques (e.g., Gaussian noise, Possion noise, pixelation, etc.).

\textbf{Defense methods}. We select several state-of-the-art adversarial defense methods, including NAT \cite{alexey2017adversarialmachine} \textcolor{black}{(adversarially training a model with FGSM using different training strategies)}, PAT \cite{madry2017towards} \textcolor{black}{(adversarial training with PGD)}, EAT \cite{tramer2017ensemble} \textcolor{black}{(adversarial training with FGSM generated by multiple models)}, LAT \cite{sankaranarayanan2018regularizing} \textcolor{black}{(injecting noise into hidden layers)} and Rand \cite{xie2018mitigating} \textcolor{black}{(randomly resizing input images)}. Among these methods, EAT and Rand were ranked No.1 and No.2 in \emph{NeurIPS 2017} adversarial defense competition.

\textcolor{black}{To ensure that our experiments are fair, we use FoolBox \cite{Rauber2017Foolbox} and select the hyper-parameters for attack and defense methods that are suggested in the relevant papers and works, e.g., \cite{alexey2017adversarialmachine,carlini2019evaluating,pmlr_v70_cisse17a,madry2017towards}, etc., which are comparable to other defense strategies. Further details can be found in the Supplementary Material.}

\subsection{Evaluation Criteria}
In this part, we will explicate the metrics used in our paper to evaluate model robustness against adversarial perturbations, corruption, and more generalized noise.

\subsubsection{Adversarial robustness evaluation} We use top-1 \emph{worst case classification accuracy} for our black-box attack defense. For a specific test set, corresponding adversarial example sets are generated using attack methods from different hold-out models. Subsequently, the worst results are selected among them as the final result. However, top-1 \emph{classification accuracy} is utilized for the white-box attack. In this situation, adversaries know every detail of the target model and generate adversarial examples for direct attack. For these evaluation metrics, the higher the better.

\subsubsection{Corruption robustness evaluation} We adopt mCE, Relative mCE and mFR, following \cite{hendrycks2018benchmarking}, to comprehensively evaluate a classifier's robustness to corruption. More specifically, mCE denotes the mean corruption error of the model compared to the baseline model, while Relative mCE represents the gap between mCE and the clean data error. Moreover, mFR stands for the classification differences between two adjacent frames in the noise sequence for a specific image. For these evaluation metrics, the lower the better.

\textbf{mCE.} The first evaluation step involves taking a classifier f, which has not been trained on CIFAR10-C, and computing the clean dataset top-1 error rate as $E^f_{clean}$. The second step involves testing the classifier on each corruption type $c$ at each level of severity $s$, denoted as $E^f_{s,c}$. \textcolor{black}{Finally, mCE is computed by dividing the errors of a baseline model as follows:}
\begin{align*}
CE_c^f = \frac{\sum_{s=1}^5 E^f_{s,c}}{\sum_{s=1}^5 E^{base}_{s,c}}.
\end{align*}
Thus, mCE is the average of 15 different Corruption Errors (CEs).

\textbf{Relative mCE.} A more nuanced corruption robustness measure is that of Relative mCE. If a classifier withstands most corruption, the gap between mCE and the clean data error is minuscule. Thus, Relative mCE is calculated as follows:
\begin{align*}
\text{Relative } mCE_c^f = \frac{\sum_{s=1}^5 E^f_{s,c} - E^f_{clean}}{\sum_{s=1}^5 E^{base}_{s,c} - E^{base}_{clean}}.
\end{align*}

\textbf{mFR.} Let us denote m noise sequences with $S=\{(x_1^{(i)},x_2^{(i)},...,x_n^{(i)})\}_{i=1}^m$, where each sequence is created with noise $p$. \textcolor{black}{The ``Flip Probability'' of network $f$ on noise sequences $S$ is:}
\begin{align*}
FP_p^f &= \frac{1}{m(n-1)} \sum_{i=1}^m\sum_{j=2}^n 1 (f(x_j^{(i)}) \neq f(x_{j-1}^{(i)}))\\& = \mathbb{P}_{x \thicksim S}((f(x_j) \neq f(x_{j-1})).
\end{align*}
%For noise sequences, which are not temporally related, $x_1^{(i)}$ is clean and $x_j^{(i)}$ $(j>1)$ are perturbed images of $x_1^{(i)}$. We can recast the FP formula for noise sequences as:
%\begin{align*}
%FP_p^f &= \frac{1}{m(n-1)} \sum_{i=1}^m\sum_{j=2}^n 1 (f(x_j^{(i)}) \neq f(x_{1}^{(i)}))\\& = %\mathbb{P}_{x \thicksim S}((f(x_j) \neq f(x_{1})\ |\ j > 1).
%\end{align*}
The Flip Rate can thus be obtained by $FR_p^f=FP_p^f/FP_p^{base}$, where mFR is the average value of FR.

\subsubsection{Model Structure Robustness}
Standard methods for both attack and defense \cite{alexey2017adversarialmachine,buckman2018thermometer,athalye2018obfuscated} typically evaluate model robustness using different variants of classification accuracy (e.g., worst-case, average-case, etc.). However this type of measurement only focuses on the final output predictions of the models (i.e., their final behaviors) and reveals only limited information regarding how and why robustness is achieved. \textcolor{black}{To further understand how model the structure of a model can affect its robustness, we evaluate model robustness from the model structure perspective. We consequently proposed two metrics, namely \emph{\textcolor{black}{Empirical} Boundary Distance} and \emph{$\varepsilon$-Empirical Noise Insensitivity}, which are based on decision boundary distance and the Lipschitz constant, respectively.}

\textbf{\textcolor{black}{Empirical} Boundary Distance}. The minimum distance to the decision boundary among the data points reflects the model robustness to small \textcolor{black}{noise} \cite{cortes1995support,NIPS2018_7364}. Due to the computation difficulty for decision boundary distance, we propose \emph{\textcolor{black}{Empirical} Boundary Distance} (denoted as $W_{f}$) in a heuristic way. Intuitively, a larger $W_{f}$ means a stronger model. Given a learnt model $f$ and point $x_i$ with class label $y_i$ ($i=1,\ldots,N$), we first generate a set $V$ of $m$ random orthogonal directions \cite{he2018decision}. For each direction in $V$, we then estimate the root mean square (RMS) distances $\phi_i(V)$ to the decision boundary of $f$ until the model's prediction changes: i.e., $f(x_i) \neq y_i$. Among $\phi_i(V)$, $d_i$ denotes the minimum distance moved to change the prediction for instance $x_i$. Our \emph{\textcolor{black}{Empirical} Boundary Distance} is thus defined as follows:
\begin{equation}
\begin{aligned}
W_{f}=\frac{1}{N}\sum_{i=1}^{N} d_i, \quad d_i= \min\phi_i(V).
\end{aligned}
\end{equation}

\textbf{$\varepsilon$-Empirical Noise Insensitivity}. \cite{xu2012robustness} first introduced the concept of learning algorithms robustness, which is based on the idea that if two samples are ``similar'', then their test errors will be very close. Inspired by this, we propose \emph{$\varepsilon$-Empirical Noise Insensitivity} to measure the model robustness against generalized noise from the view of the Lipschitz constant. Evidently, a lower value indicates a stronger model. We first randomly select $N$ clean examples, after which $M$ examples are generated from each clean example via various methods (e.g., adversarial attack, Gaussian noise, blur, etc.). The differences between model loss functions are computed when a clean example and corresponding polluted examples are fed into the model. The different severities in the loss function are used to measure the model's insensitivity and stability to generalized small \textcolor{black}{noise} within constraint $\varepsilon$:
\begin{equation}
\begin{aligned}
I_f(\varepsilon)=&\frac{1}{N \times M}\sum_{i=1}^{N} \sum_{j=1}^{M} \frac{|l_f(x_i|y_{i})-l_f(\mu_{ij}|y_{i})|}{\|x_i-\mu_{ij}\|_\infty}\\
&s.t. \quad \|x_i-\mu_{ij}\|_\infty \leq \varepsilon,
\end{aligned}
\end{equation}
where $x_i$, $\mu_{ij}$ and $y_i$ denote the clean example, corresponding polluted example and class label, respectively. $l_f(\cdot|\cdot)$ represents the loss function of model $f$.

%\subsubsection{Implementation Details}
%In our experiments, we use Pytorch for the implementation and test them on a NVIDIA Tesla K80 GPU cluster. All the models for MNIST, CIFAR-10 and ImageNet are trained for 40, 140 and 100 epochs, respectively. As for the attack and defense methods, we implement them following their original papers. More detailed information of training and hyper-parameters can be found in the supplementary material.

\subsection{Is It Necessary to Inject \textcolor{black}{Noise} into All Layers?}

\begin{figure}[!htb]
\centering
%\vspace{-0.15in}
%\hspace{-0.2in}
\subfigure[top-m layer group]{
\includegraphics[width=0.46\linewidth]{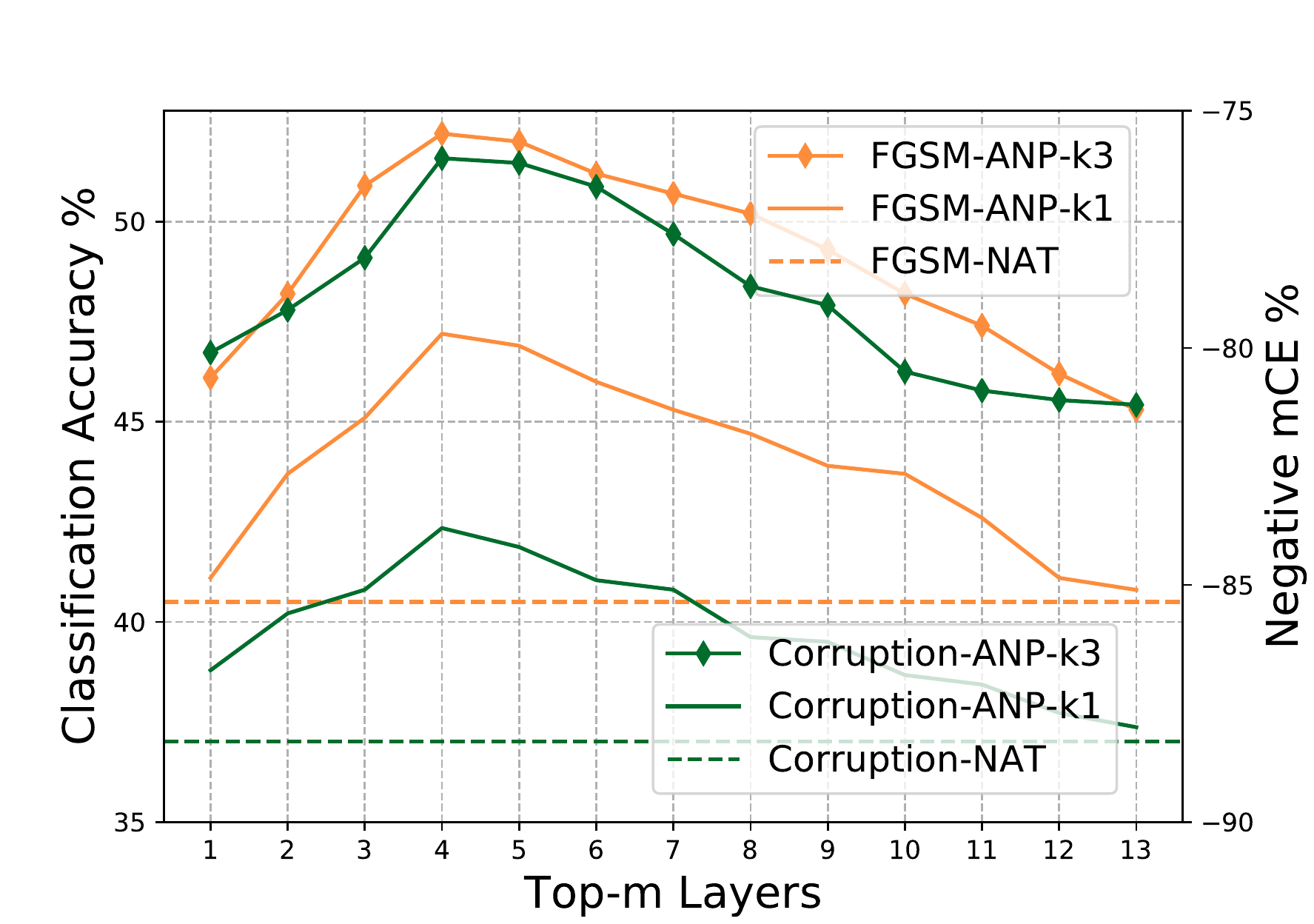}
}
\subfigure[bottom-m layer group]{
\includegraphics[width=0.46\linewidth]{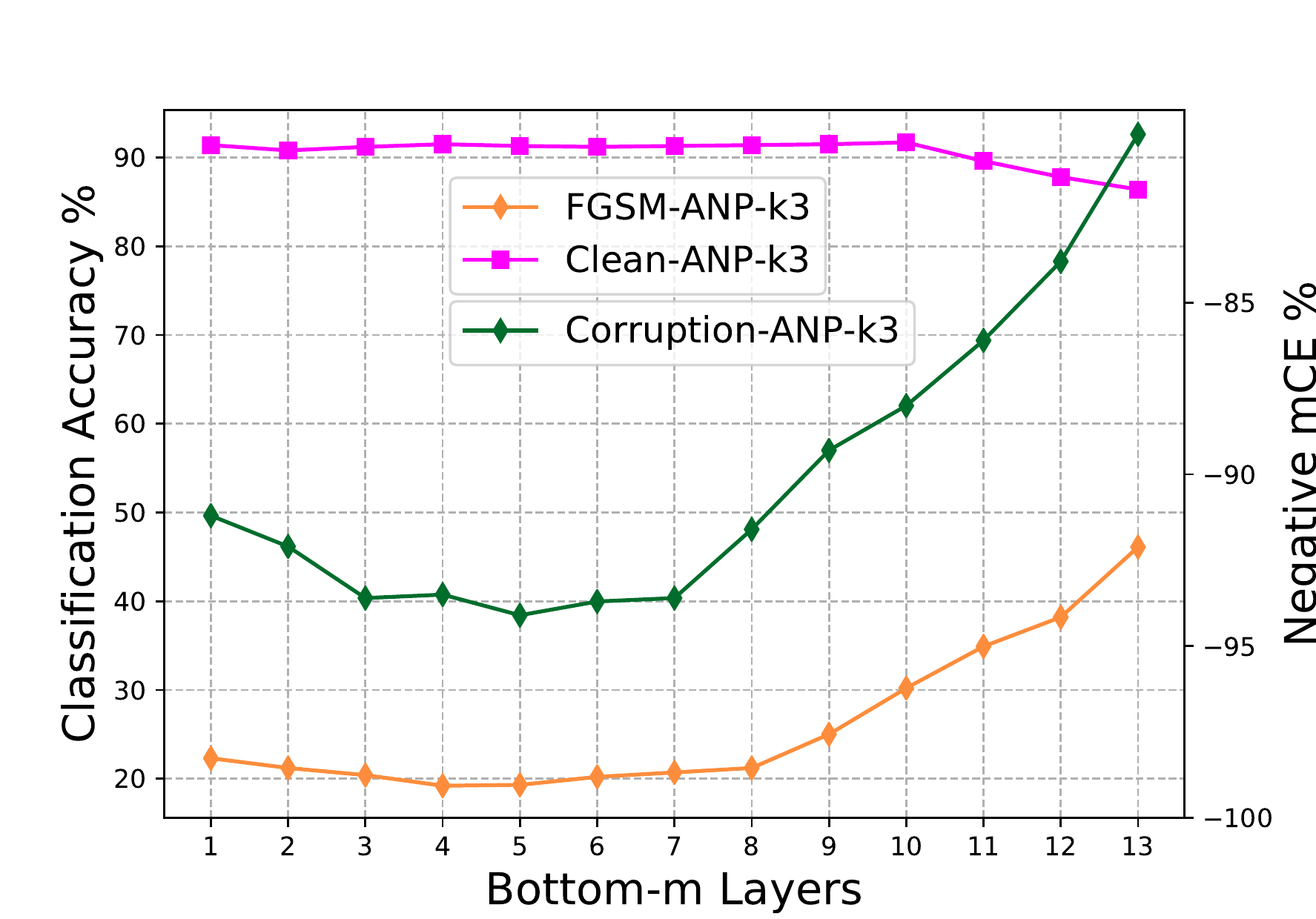}
}

%\vspace{-0.1in}
\caption{\textcolor{black}{The results of VGG-16 trained with adversarial noise added to different top-m and bottom-m layer groups. The horizontal axis of (b) is arranged in an order opposite to that of the layer number (i.e., by adding \textcolor{black}{noise} to the bottom-1 layers, we mean perturbing the 13th layer). The solid lines denote models trained with different layer groups via ANP, while the dashed lines represent the baseline method NAT. The model achieves the best performance when noise is injected only into the top-4 layers.}}
\label{fig:layergroup}
%\vspace{-0.1in}
\end{figure}

\begin{figure}[!htb]
\centering
%\vspace{-0.15in}
%\hspace{-0.2in}
%\subfigure[]{
%\includegraphics[width=0.86\linewidth]{res/layergroup.pdf}
%}
%\hspace{-0.15in}
\subfigure[1st layer]{
\includegraphics[width=0.46\linewidth]{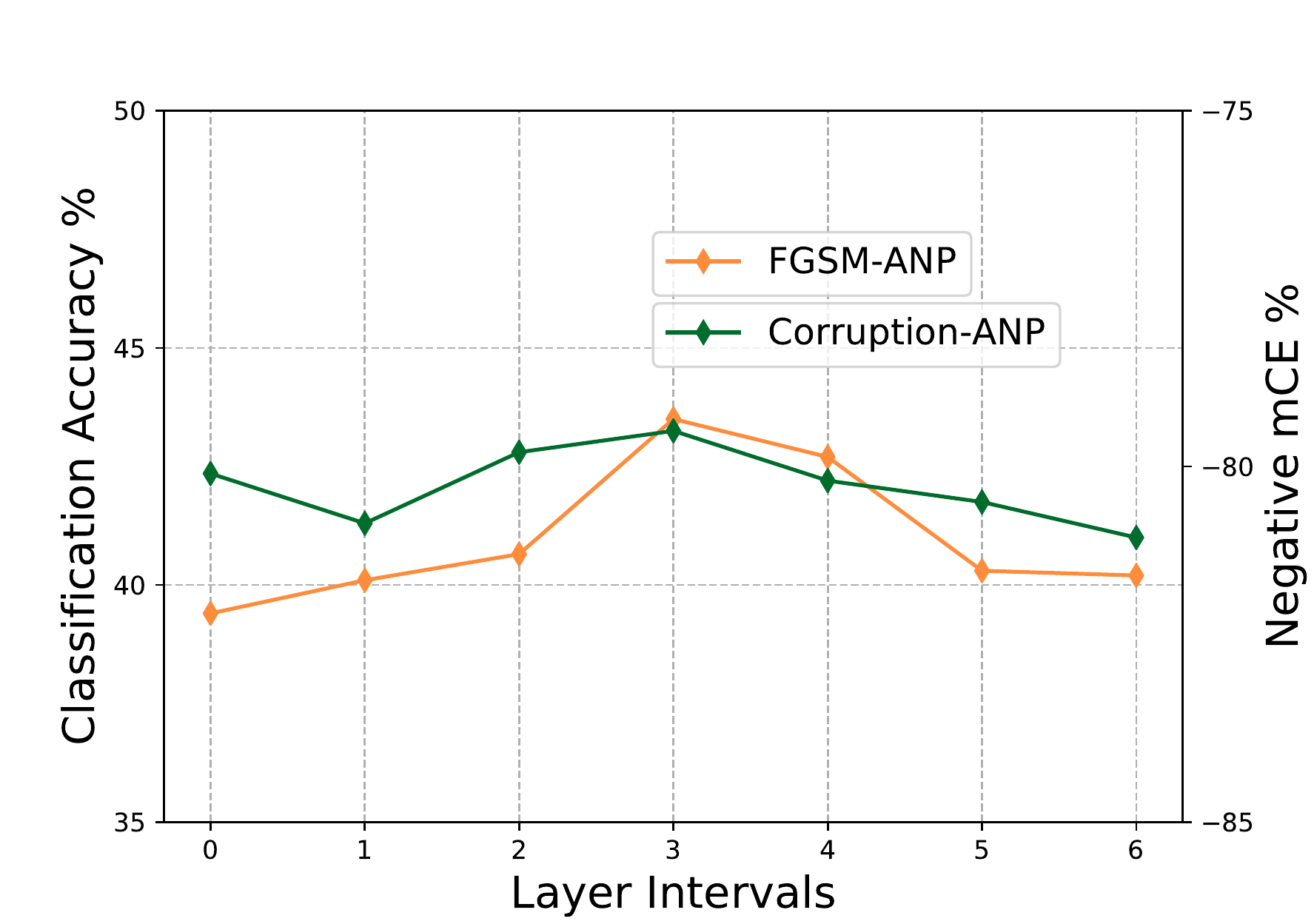}
}
\subfigure[7th layer]{
\includegraphics[width=0.46\linewidth]{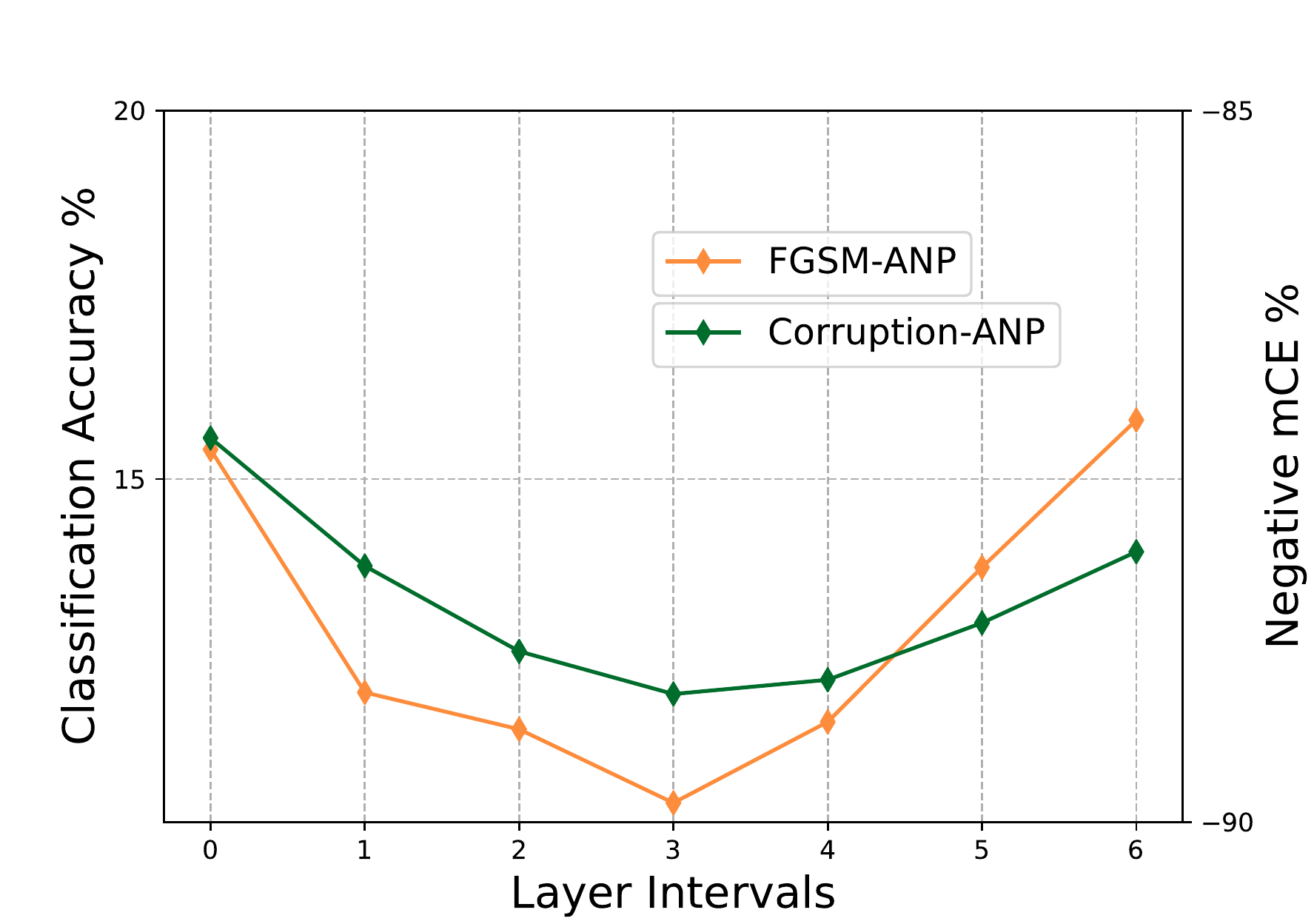}
}
%\vspace{-0.1in}
\caption{\textcolor{black}{The results of VGG-16 trained with adversarial noise added to distant and neighboring layers. In subfigure (a) and (b), We select the 1st and 7th layer respectively as the base layer. We perturb a pair of layers with different intervals ranging from 0 to 6.}}
\label{fig:distant_neighbor}
%\vspace{-0.1in}
\end{figure}

\begin{figure}[!htb]
\centering
%\vspace{-0.15in}
%\hspace{-0.2in}
%\subfigure[]{
%\includegraphics[width=0.86\linewidth]{res/layergroup.pdf}
%}
%\hspace{-0.15in}
\subfigure[]{
\includegraphics[width=0.46\linewidth]{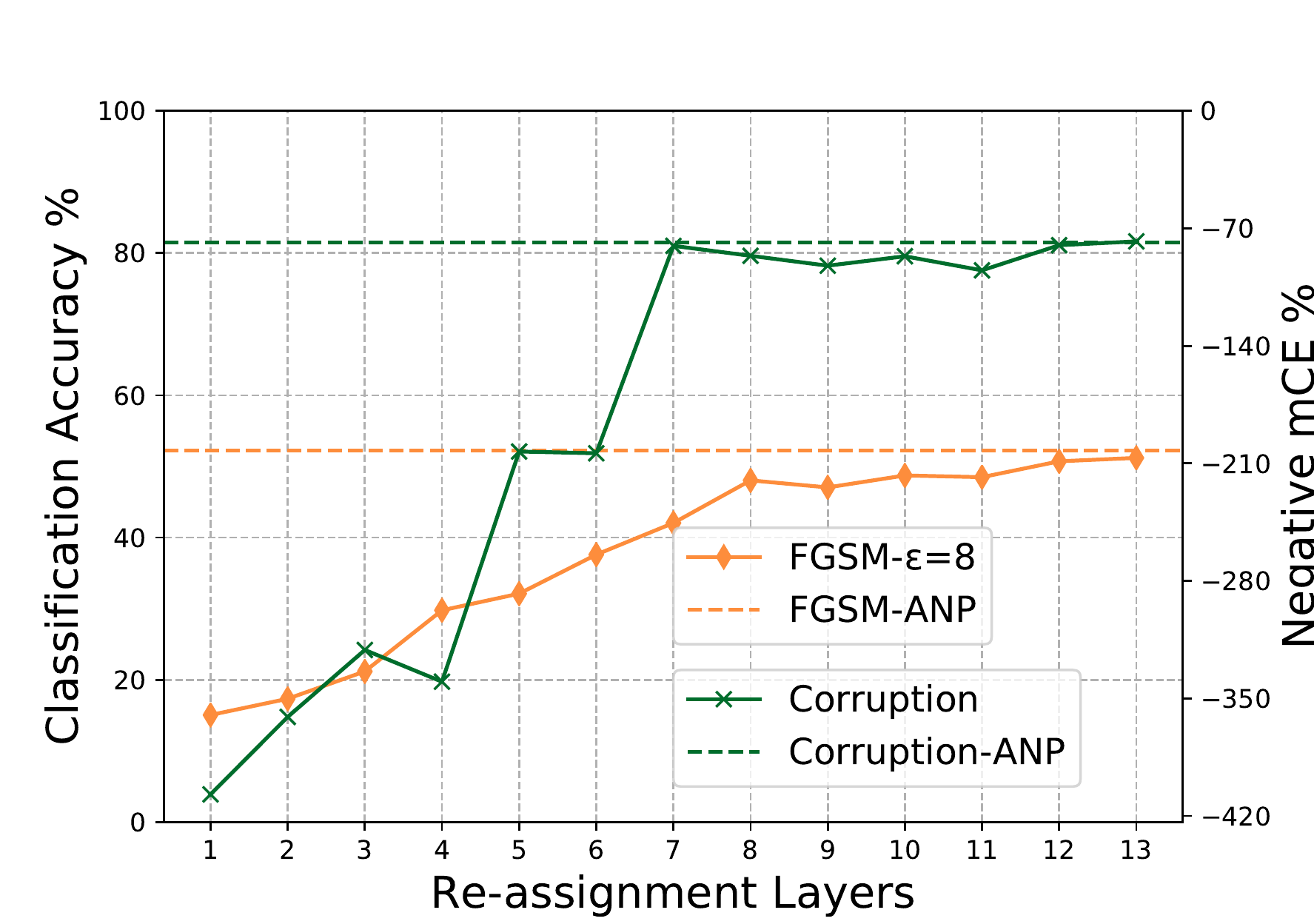}
}
\subfigure[]{
\includegraphics[width=0.46\linewidth]{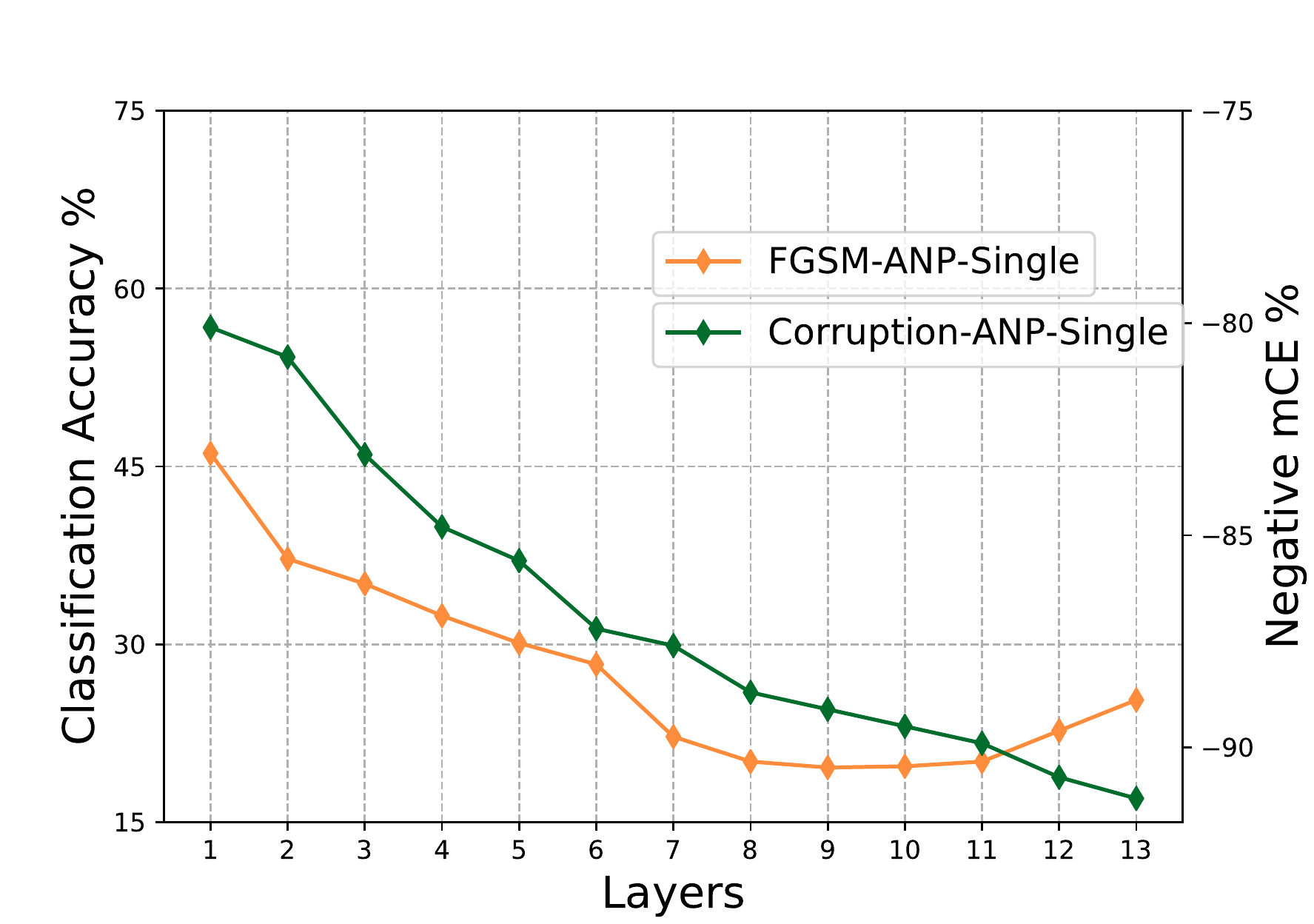}
}
%\vspace{-0.1in}
\caption{Subfigure (a) denotes the layer statistics re-assignment experiment; here dashed lines represent the baseline model with no layer weight re-assignment, i.e. ANP. Subfigure (b) illustrates single hidden layer perturbing experiment, i.e., in which we only perturb the $m$-th layer. The model robustness decreases as the layer goes deeper.}
\label{fig:transplant}
%\vspace{-0.1in}
\end{figure}

A significant body of research exists regarding the representation power of the neural network \cite{hornik1991approximation,delalleau2011shallow}; however, studies for hidden layers are rarely involved. Since not all layers are created equal \cite{zhang2019all}, an intuitive question for ANP training emerges, namely: \emph{do we need to add \textcolor{black}{noise} to all layers?} We therefore try to investigate the contribution of hidden layers when adversarial \textcolor{black}{noise is} injected and propagated. \textcolor{black}{In this section, all adversarial examples are generated via white-box FGSM with $\varepsilon$=8 in terms of $\ell_{\infty}$ norm.}

\textcolor{black}{\subsubsection{Layer group study}}
\textcolor{black}{We first study the contribution made by hidden layers from the perspective of layer groups, i.e., multiple layers.} First, we train a number of VGG-16 models on CIFAR-10 with adversarial \textcolor{black}{noise} injected into different layer groups: namely, the top-$m$ \textcolor{black}{and the bottom-$m$} layers. From the results in Figure \ref{fig:layergroup} (a), we can observe that the model robustness improves (i.e., both the classification accuracy for adversarial examples and the negative mCE for corruption increase) when injecting \textcolor{black}{noise} into the top-$m$ ($m\leq 4$) hidden layers. \textcolor{black}{We can further observe a similar phenomenon from Figure \ref{fig:layergroup} (b): as we increase the number of bottom layers to be perturbed, the model robustness increases, especially when the shallow layers are involved. Surprisingly, however, it is not true that the model becomes increasingly robust the more layers are perturbed;} this indicates that different hidden layers contribute to different extents to model robustness in deep architecture. \textcolor{black}{Shallow} layers are more critical to model robustness; by contrast, the importance of \textcolor{black}{deep} layers is somewhat more limited.

Meanwhile, we trained ANP with the progressive number $k$=1; this is roughly similar to NAT with \textcolor{black}{noise} confined only to the top-1 layer. Its performance curves are indicated by the orange and green solid lines without a marker (ANP-k1) in Figure \ref{fig:layergroup} (a), which are almost close to that of NAT, while much lower than the model trained with $k=3$ (ANP-k3). This, in turn, demonstrates the importance of our progressive noise injection to model robustness.

\textcolor{black}{We further investigate the contributions of different layers by perturbing both neighboring layers and distant layers. For a VGG-16 model, we choose to add \textcolor{black}{noise} to a pair of layers by selecting a base layer and second layer, with different intervals between these two layers ranging from 0 to 12. In other words, we add \textcolor{black}{noise} to neighboring layers when the interval is 1, and perturb more distant layers when interval exceedsan 1. As shown in Figure \ref{fig:distant_neighbor}, adding distant layers at specific positions is superior to adding \textcolor{black}{noise} to two neighboring layers; for example, the model is most robust when we perturb the 1st and 4th layer in Figure \ref{fig:distant_neighbor} (a), or the 7th and 13th layer in Figure \ref{fig:distant_neighbor} (b). We attribute these results to the strong contributions made by the shallow layers towards model robustness, since we also obtain the strongest model when we choose the top-4 layers in Fig 2 (a).}

\textcolor{black}{\subsubsection{Single-layer study}}
We further investigate the contributions made by hidden layers \textcolor{black}{from the perspective of single layer} via layer statistics re-assignment and single hidden layer perturbation.

\textcolor{black}{We first re-assign the statistics of each Batch Normalization (BN) \cite{ioffe2015batch} layer (i.e., running mean and running variance) of an ANP-trained VGG16 model using a corresponding vanilla model at the time of inference. Intuitively speaking, a stronger model performance gap following layer statistics re-assignment results in larger layer-wise differences, which in turn indicate a more non-trivial hidden layer on model robustness. As shown in Figure \ref{fig:transplant} (a), the influence of individual layer statistics re-assignment to model robustness reduces (i.e., the performance gap is smaller) as the layer depth increases.}

We then further perturb single hidden layers individually; i.e., \textcolor{black}{noise} is injected only into the $m$-th layer. As shown in Figure \ref{fig:transplant} (b), the model robustness reduces (i.e., classification accuracy for adversarial examples and negative mCE for corruptions reduces) as the layer depth increases.

Thus, the experiments above constitute double confirmation that \textcolor{black}{shallow} layers are more critical to model robustness, while on the contrary, the importance of \textcolor{black}{deeper} layers is somewhat lesser.

{\color{black}{\subsubsection{Theoretical analysis}
In this section, we try to provide theoretical analysis for the above conclusions (model performance with or without ANP).

We first consider the feed-forward neural networks with ReLU activation. Here, the function $F$ is parameterized by a sequence of matrices $W = (W_1, W_2, \ldots, W_L)$, i.e., $F = F_W$. For the sake of convenience, we assume that $W_h \in \mathbb{R}^{d_h \times d_{h-1}}$, where $d_h$ and $d_{h-1}$ are the dimensions of two adjacent layers, while $\rho(\cdot)$ is the ReLU function. For the vectors, $\rho(x)$ is the vector generated by applying $\rho(\cdot)$ on each coordinate of $x$, i.e., $\rho(x) ]_i = \rho(x_i)$. Thus, we have
$$ F_W (x) = W_L \rho ( W_{L-1} \rho (\cdots  \rho(W_{1}x)  \cdots)  ). $$

For a $K$-classification problem, we have the dimension $d_L = K$, $F_W (x) : \mathbb{R}^d \rightarrow \mathbb{R}^K$, while $[F_W (x)]_k$ is the score for the $k$-th class.

Let $\epsilon_l \in \{\epsilon_1,\epsilon_2,\cdots,\epsilon_L\}$ denote the adversarial perturbation added in the hidden layers of ANP and corresponding perturbation vector $\bm{\epsilon_l}=\epsilon_l \cdot \bm{1_{d_l}}$ with dimension $d_l$ in the $l$-th layer. We denote the function $f$ with layer-wise adversarial perturbations of ANP as follows:
$$ F_{W,\epsilon} (x) = W_L( \rho ( W_{L-1}( \rho (\cdots  \rho(W_{1}(x+\bm{\epsilon_1}))  \cdots) + \bm{\epsilon_{L-1}})) + \bm{\epsilon_L}). $$

\begin{theorem} \label{theorem 2}
 Consider the function $F_W(\cdot)$ and the layer-wise adversarial perturbation vector $\bm{\epsilon_l}$ in the $l$-th layer. Given an input sample $x$, the difference between the $F_{W,\epsilon} (x)$ (model trained with ANP) and $F_W(x)$ (model trained without ANP, i.e., vanilla) could be derived as
 \begin{equation*} %\label{eqa}
 \begin{split}
  \|F_{W,\epsilon}& (x) - F_W (x)\| \leq \\
  & \|\Pi_{i=1}^L W_i \bm{\epsilon_1}\| + \|\Pi_{j=2}^{L-1} W_j \bm{\epsilon_{L-1}}\| + \cdots + \|W_L \bm{\epsilon_L}\|
  \end{split}
 \end{equation*}
\end{theorem}
Consequently, the difference in output scores (expressed in the accuracy on clean sample $x$) between classifiers $F_{W,\epsilon} (x)$ and $F_W (x)$ could be influenced more strongly by the adversarial perturbations in shallow layers. As can be seen from Figure \ref{fig:layergroup} (b), as we perturb more bottom-m layers (from deep layers to shallow layers), the accuracy on clean samples is decreasing. Thus, we can draw the following conclusions: (1) shallow layers have stronger negative influences than deep layers on model accuracy for clean examples; (2) shallow layers are more critical to model robustness than deep layers; (3) the contribution of each layer with respect to layer depth is not strictly linear.}}

In summary, it is sufficient to \emph{inject noise just the \textcolor{black}{shallow} layers only to achieve better model robustness}; this finding will be used to guide our experiments throughout the rest of the paper.

%which double confirms the above conclusion. Lastly, we study the performance of ANP on models with different depths, i.e., numbers of hidden layers. Experimental results indicate that models trained with ANP shows barely superiority between shallow and deep architectures. Detailed information can be found in the supplementary material.

%We also conduct the experiment for randomly perturbing each layer with gaussian noise N(0, 0.1), see below for the brief results on CIFAR-10 with VGG-16, (clean, FGSM, mCE): 88.5\%, 51.3\%, 118.5\%, which is worse than ANP.

\begin{table*}[!thb]
%\vspace{-0.15in}
\caption{Black-box and white-box attack defense results on MNIST with LeNet.}
\label{tab:mnist}
\begin{center}
\begin{small}
\begin{sc}
\begin{tabular}{lccccccc}
\toprule
 &  & \multicolumn{3}{c}{{Black-box}} & \multicolumn{3}{c}{{White-box}}\\
\cline{3-8}
LeNet & clean & \multicolumn{3}{c|}{{FGSM}} & \textcolor{black}{BIM} & PGD & C\&W\\
 & & \scriptsize $\varepsilon$=0.1 & \scriptsize $\varepsilon$=0.2 & \multicolumn{1}{c|}{\scriptsize $\varepsilon$=0.3}& {\scriptsize $\varepsilon$=0.2} & {\scriptsize $\varepsilon$=0.2} &   \\
%\midrule
\hline
Vanilla    & \textbf{99.6\%} & 72.0\% & 28.0\% & \multicolumn{1}{c|}{4.0\%}& 61.7\% & 22.3\% & 27.1\% \\
PAT & 99.0\% & 96.8\% & 90.7\% & \multicolumn{1}{c|}{\textbf{78.0\%}} & 90.6\% & \textbf{59.1\%} & 63.7\% \\
NAT    & 98.4\% & 94.0\% & 88.0\% & \multicolumn{1}{c|}{72.0\%} & 85.9\% & 45.1\% & 51.4\% \\
\textbf{ANP}  & 99.3\% & \textbf{97.1\%} & \textbf{93.2\%} & \multicolumn{1}{c|}{\textbf{78.0\%}}& \textbf{91.5\%} & \textbf{59.1\%}& \textbf{69.1\%} \\
\bottomrule
\end{tabular}
\end{sc}
\end{small}
\end{center}
\end{table*}

\begin{table*}[!thb]
%\vspace{-0.1in}
\caption{Black-box attack defense results on CIFAR-10 with different models.}
\label{tab:tab2}
\begin{center}
\begin{small}
\begin{sc}
\setlength{\tabcolsep}{2.5mm}{
\subtable[VGG-16]{
\begin{tabular}{ccccccc}
\toprule
VGG-16 & clean & \multicolumn{2}{c}{{FGSM}} & PGD & Step-LL & MI-FGSM  \\
\cline{3-7}
 & & \scriptsize $\varepsilon=8$& \scriptsize $\varepsilon=16$ & {\scriptsize $\varepsilon=8$, $\alpha=2$}
 & {\scriptsize $\varepsilon=8$} & {\scriptsize $\varepsilon=8$}     \\
 \midrule
 Vanilla & \textbf{92.1\%} & 38.4\% & 19.3\% & 0.0\% & 7.5\% &2.3\% \\
 PAT & 83.1\% & 82.5\% & 76.3\% & \textbf{85.5\%} & 80.1\% & 79.9\% \\
 NAT & 86.1\% & 73.5\% & 70.2\% & 80.3\% & 79.1\% &77.6\% \\
 LAT & 84.4\% & 75.8\% & 63.7\% & 79.2\% & 78.3\% &77.6\% \\
 Rand & 85.2\% & 77.6\% & 70.8\% & 80.2\% & 79.3\% &78.2\% \\
 EAT & 87.5\% & 81.2\% & 76.2\% & 83.5\% & 82.7\% &80.8\% \\
 \textbf{ANP} & 91.7\% & \textbf{82.8\%} & \textbf{76.4\%} & 84.4\% & \textbf{83.3\%} &\textbf{81.1\%} \\
\bottomrule
\end{tabular}}

\subtable[ResNet-18]{
\begin{tabular}{ccccccc}
\toprule
ResNet-18 & clean & \multicolumn{2}{c}{{FGSM}} & PGD & Step-LL & MI-FGSM   \\
\cline{3-7}
 & & \scriptsize $\varepsilon=8$ & \scriptsize $\varepsilon=16$ & \scriptsize $\varepsilon=8$, $\alpha=2$
 & \scriptsize $\varepsilon=8$ & \scriptsize $\varepsilon=8$ \\
 \midrule
 Vanilla & \textbf{93.1\%} & 12.8\% & 10.2\% & 6.0\% & 21.4\% &1.0\% \\
 PAT & 85.1\% & 82.4\% & 72.4\% &
 \textbf{87.4\%} & 81.0\% & 79.8\% \\
 NAT & 89.1\% & 78.1\% & 68.8\% & 83.6\% & 80.4\% &77.7\% \\
 LAT & 88.9\% & 45.8\% & 23.1\% & 58.3\% & 49.7\% &33.0\% \\
 Rand & 86.4\% & 57.6\% & 35.0\% & 70.4\% & 61.0\% &33.0\% \\
 EAT & 86.9\% & 80.8\% & 72.5\% & 84.7\% & 82.8\% &80.4\% \\
 \textbf{ANP} & 92.1\% & \textbf{83.6\%} & \textbf{73.5}\% & 86.5\% & \textbf{84.0\%} &\textbf{80.4\%} \\
\bottomrule
\end{tabular}

}
}
\end{sc}
\end{small}
\end{center}
%\vspace{-0.25in}
\end{table*}

\subsection{Adversarial Robustness Evaluation}
We first evaluate model adversarial robustness by considering black-box and white-box attack defense.

\subsubsection{Black-box setting}
In black-box attack defense, \textcolor{black}{adversaries have no knowledge of the target models} (e.g., architectures, parameter weights, etc.). We first generate adversarial examples (10k images) using various hold-out models which are different from the target models; we then use these adversarial examples to attack the target model.

\textbf{MNIST.} With LeNet as the target model, the experimental results for black-box and white-box defense are listed in Table \ref{tab:mnist}.

\textbf{CIFAR-10.} As shown in Table \ref{tab:tab2}, on CIFAR-10, we employ VGG-16 as the target model and compare ANP with various other defense methods. The hold-out models include ResNet-50, DenseNet and Inception-v2. Among these defense methods, EAT is trained ensemble with VGG-16 and Inception-v2; meanwhile, Rand resizes input images from 32 to 36 and follows a NAT-trained VGG-16.
%Due to the limited page, experiments with target model ResNet-18 are presented in the supplementary material.

\textbf{ImageNet.} AlexNet is applied as the target model, and adversarial examples are generated from ResNet-18 and AlexNet. Meanwhile, EAT is trained with AlexNet and ResNet-18, while Rand resizes input images from 224 to 254 and follows a NAT-trained AlexNet. The results on ImageNet are presented in Table \ref{tab:tab5}.

\begin{table*}[!thb]
\caption{Black-box attack defense results on ImageNet with AlexNet.}
\label{tab:tab5}
%\vspace{-0.1in}
\begin{center}
\begin{small}
\begin{sc}
\setlength{\tabcolsep}{2.5mm}{
\begin{tabular}{ccccccc}
\toprule
AlexNet & clean & \multicolumn{2}{c}{{FGSM}} & PGD  & Step-LL & MI-FGSM  \\
\cline{3-7}
  & & \scriptsize $\varepsilon=8$& \scriptsize $\varepsilon=16$  & {\scriptsize $\varepsilon=8$, $\alpha=2$}
 & {\scriptsize $\varepsilon=8$} & {\scriptsize $\varepsilon=8$}   \\
 \midrule
 Vanilla & \textbf{61.7\%} & 12.6\% & 9.2\%  & 4.3\% & 13.7\% &3.5\%  \\
 PAT & 56.2\% & 41.5\% & \textbf{40.2\%} & \textbf{42.1\%} & 41.5\% & \textbf{41.2\%} \\
 NAT & 53.5\% & 39.1\% & 34.2\%  & 39.6\% & 41.6\% &37.7\% \\
 RAND & 50.2\% & 39.2\% & 27.0\% & 39.2\% & 40.9\% &39.7\%  \\
 EAT & 51.1\% & 39.6\% & 35.2\%  & 39.9\% & 42.3\% &37.9\% \\
 \textbf{ANP} & 51.5\% & \textbf{41.7\%} & 39.3\% & \textbf{42.1\%} & \textbf{42.7\%} &\textbf{41.2\%}\\
\bottomrule
\end{tabular}}
\end{sc}
\end{small}
\end{center}
%\vspace{-0.1in}
\end{table*}
From the above black-box experiments, we can make the following observations:
(1) In the black-box setting, in almost all cases, ANP achieves the best defense performance among all methods; (2) ANP makes the widely used deep models strongly robust against both single-step and iterative black-box attacks; (3) Normally, the classification performance for clean examples degrades significantly (e.g., by 5-10\% in terms of accuracy) when more noise is introduced into the model; however, ANP supplies the models with good generalization ability, resulting in stable classification accuracy that is close to that of the vanilla models on CIFAR-10 as well as MNIST.

\begin{table}[!thb]
%\vspace{-0.1in}
\caption{White-box attack defense on CIFAR-10 and ImageNet.}
\label{tab:tab3}
\vspace{-0.15in}
\begin{center}
\begin{small}
\begin{sc}
%\hspace{-0.1in}
\subtable[CIFAR-10 with VGG-16]{
\begin{tabular}{ccccc}
\toprule
VGG-16 & clean & BPDA & PGD & C\&W \\
\cline{3-5}
 & & {\scriptsize $\varepsilon$=8} & {\scriptsize $\varepsilon$=8} & \\
\midrule
Vanilla    & \textbf{92.1\%} & 0.2\% & 0.0\%& 8.2\% \\
PAT    & 83.1\% & 41.9\% & \textbf{41.4\%} & 42.4\% \\
NAT    & 86.1\% & 24.5\% & 8.1\% &31.6\%\\
%LAT    & 84.4\% & 25.5\% & 17.6\% \\
Rand    & 85.2\% & 0.2\% & 9.1\% &34.2\%\\
EAT    & 87.5\% & 37.6\% & 9.9\% & 35.2\%\\
\textcolor{black}{Free-4}    & \textcolor{black}{88.9\%} & \textcolor{black}{41.2\%} & \textcolor{black}{38.2\%} & \textcolor{black}{42.3\%}\\
\textbf{ANP}  & 91.7\% & \textbf{43.5\%} & 28.9\% & \textbf{48.1\%} \\
%\textcolor{black}{\textbf{ANP+PAT}}  & \textcolor{black}{92.1}\% & \textcolor{black}{\textbf{54.0\%}} & \textcolor{black}{\textbf{29.0\%}} & \textcolor{black}{\textbf{47.3\%}} \\
\bottomrule
\end{tabular}
}
\subtable[CIFAR-10 with ResNet-18]{
\begin{tabular}{ccccc}
\toprule
ResNet-18 & clean & BPDA & PGD & C\&W\\
\cline{3-5}
 & & \scriptsize $\varepsilon$=8 & \scriptsize $\varepsilon$=8 & \\
\midrule
Vanilla    & \textbf{93.1\%} & 0.0\% & 2.7\% & 8.4\%\\
PAT    & 85.1\% & 45.0\% &  \textbf{41.9\%} & 43.9\%\\
NAT    & 89.1\% & 33.5\% & 10.1\% & 32.6\%\\
Rand    & 86.4\% & 1.9\% & 9.4\% & 31.2\%\\
EAT    & 86.9\% & 40.1\% & 14.5\% & 39.1\%\\
\textcolor{black}{Free-4}    & \textcolor{black}{89.0\%} & \textcolor{black}{42.9\%} & \textcolor{black}{37.8\%} & \textcolor{black}{44.1\%}\\
\textbf{ANP}  & 92.1\% & \textbf{54.0\%} & 29.9\% & \textbf{47.3\%} \\
%\textcolor{black}{\textbf{ANP+PAT}}  & \textcolor{black}{92.1\%} & \textcolor{black}{\textbf{54.0\%}} & \textcolor{black}{\textbf{29.0\%}} & \textcolor{black}{\textbf{47.3\%}} \\
\bottomrule
\end{tabular}
}
%\hspace{-0.0in}
\subtable[ImageNet with AlexNet]{
\begin{tabular}{cccc}
\toprule
AlexNet & CLEAN & BPDA & PGD \\
\cline{3-4}
 & & {\scriptsize $\varepsilon$=8} & {\scriptsize $\varepsilon$=8} \\
\midrule
Vanilla    & \textbf{61.7\%} & 7.9\% &  2.4\% \\
PAT    & 56.2\% & 27.6\% & \textbf{28.7\%} \\
NAT    & 53.3\% & 26.8\% & 15.2\% \\
Rand    & 50.2\% & 21.6\% & 16.7\% \\
EAT    & 56.0\% & 25.0\% & 12.5\% \\
\textbf{ANP}  & 53.5\% & \textbf{28.2\%} & 27.4\% \\
\bottomrule
\end{tabular}
}
\end{sc}
\end{small}
\end{center}
%\vspace{-0.15in}
\end{table}

\subsubsection{White-box setting}
In the white-box scenario, we apply PGD, C\&W and the attacking framework BPDA for adversarial attacks. \textcolor{black}{Since we compare with Rand, which is an adversarial defense strategy that obfuscates gradients by resizing and padding the input images, we also use BPDA to circumvent the undifferentiable compnonent and ensure a thorough analysis.} Moreover, BIM is utilized in BPDA to demonstrate the white-box attack after circumventing the gradient mask with iteration number 5. The results on CIFAR-10 and ImageNet are listed in Table \ref{tab:tab3}.

\textcolor{black}{We can conclude that in the white-box setting, ANP exhibits a significant advantage in terms of defending against adversarial examples over other methods on these datasets, although ANP is slightly weak against PGD attack on ImageNet compared to PAT. Overall, the results indicate that training with ANP enables the model to be robust against various attack methods.} However, we can also see that for large datasets like ImageNet, clean accuracy drops for all methods. The reasons for this might be two-folded. Firstly, AlexNet, a relatively small model, may not have sufficient capacity to fit adversarial \textcolor{black}{noise} while maintaining high accuracy on clean examples. Secondly, the trade-off between robustness and accuracy does exist \cite{tsipras2018robustness,zhang2019theoretically} especially for high-dimensional data distributions. We will study this in more depth in future work.

\subsection{Corruption Robustness Evaluation}\label{subsec:corr}

\begin{figure}[!htb]
\centering
%\vspace{-0.15in}
%\hspace{-0.2in}
\subfigure[CIFAR-10-C]{
\includegraphics[width=0.46\linewidth]{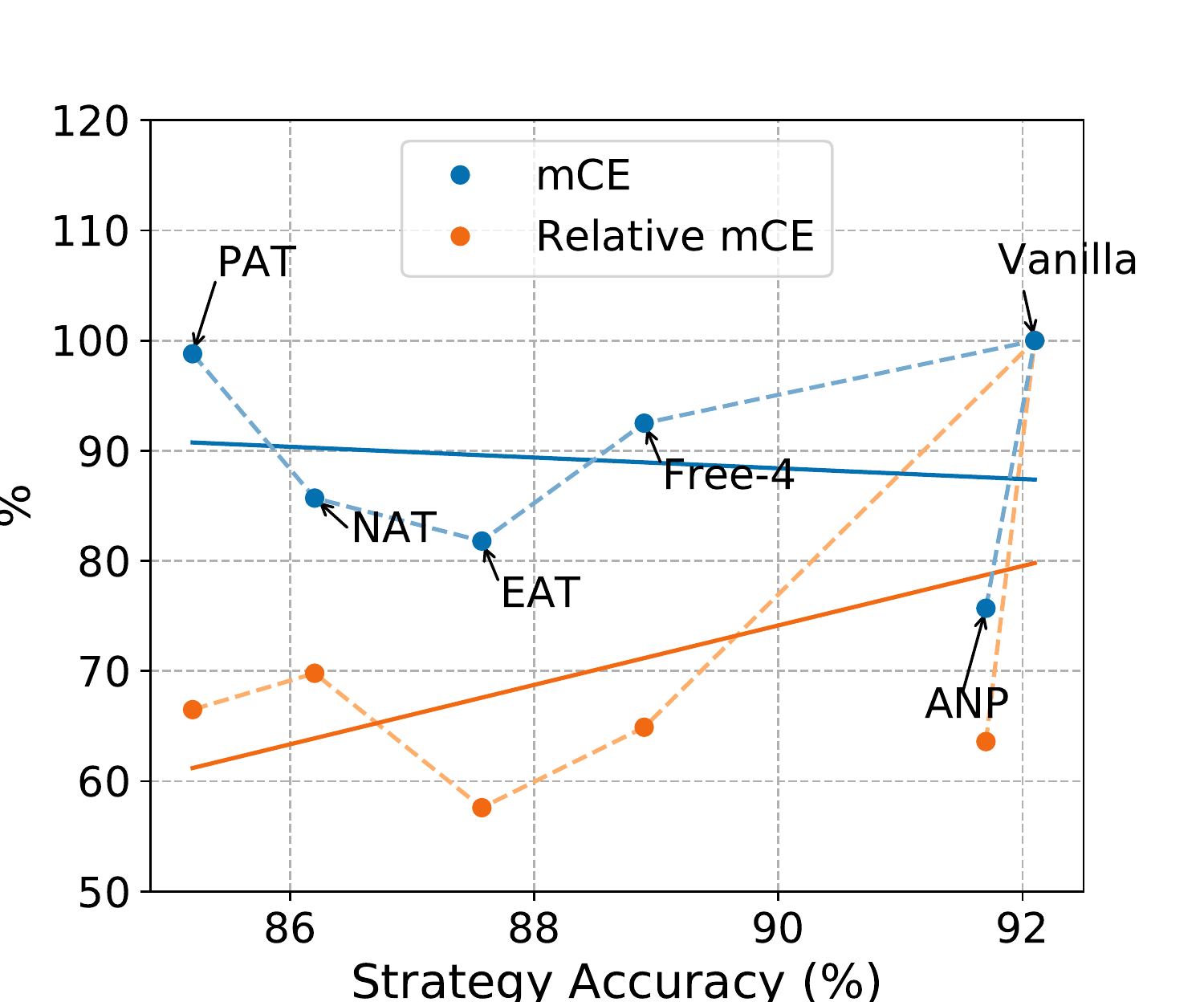}
}
%\hspace{-0.15in}
\subfigure[CIFAR-10-P]{
\includegraphics[width=0.46\linewidth]{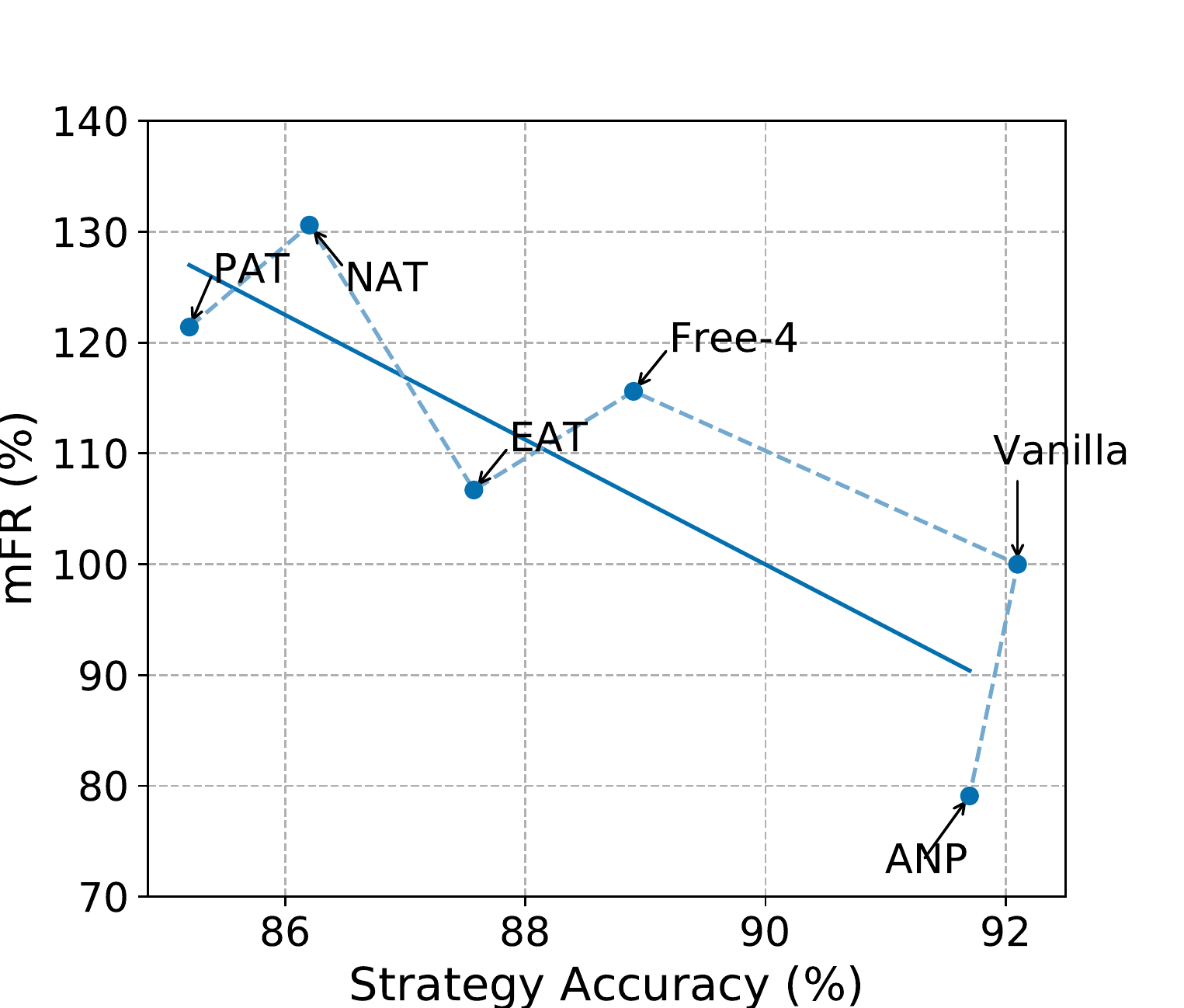}
}
%\vspace{-0.1in}
\caption{\textcolor{black}{Model corruption robustness evaluation. Our ANP outperforms other compared methods by large margins in terms of corruption robustness, with lower mCE and mFR values.}}
\label{fig:CIFAR10-CP}
%\vspace{-0.1in}
\end{figure}

To assess corruption robustness, we conduct experiments using 10K images from CIFAR-10-C with 15 different corruption levels and 5 severity levels. To test the model's dynamic robustness, we use CIFAR-10-P, which differs from CIFAR-10-C in that noise sequences are generated for each image with more than 30 frames. As shown in Section 5.2.2, mCE indicates the average corruption error, while mFR is the average flip rate of noise sequence (for both of these, lower is better). According to the results in Figure \ref{fig:CIFAR10-CP} (a) and (b), ANP achieves the lowest mCE and mFR value among all methods, indicating strong corruption robustness. More precisely, as can be seen from Table \ref{tab:detailcor}, ANP surpasses the compared strategies by large margins (i.e., almost 6 and 30 for mCE and mFR, respectively). The results demonstrate that ANP can reliably provide both static and dynamic robustness against corruption.

Although they perform well on adversarial examples, compared methods (especially PAT) show weak robustness to both static and dynamic corruptions. Another interesting phenomenon that can be observed is that all compared methods even perform worse than the vanilla model for dynamic corruption (as shown by higher mFR values). Most adversarial training methods attempt to inject \textcolor{black}{noise} into the inputs by searching for the worst-case perturbations, which indeed improve adversarial model robustness. However, these tactics seem to be worthless to average-case or general perturbations, and maybe also somehow counteract corruption robustness. Since robustness requires high data complexity \cite{schmidt2018adversarially}, our proposed ANP introduces adversarial \textcolor{black}{noise} with high complexity and diversity via progressive iteration, which contributes both adversarial and corruption robustness. We can therefore conclude that ANP supplies models with stronger corruption robustness compared to other defense methods.

We also test a VGG-16 model trained with Gaussian noise $N(0,0.1)$ added to input; however the model is weak to corruption with an mCE value 117.2.

\begin{table}[!thb]
%\vspace{-0.1in}
\caption{Corruption robustness evaluation with mCE and mFR.}
\label{tab:detailcor}
\begin{center}
\begin{small}
\begin{sc}
\setlength{\tabcolsep}{2.5mm}{
\begin{tabular}{cccc}
\toprule
VGG-16 & Error & mCE  & mFR \\
 \midrule
 Vanilla & \textbf{7.9} & 100.0 &  100.0 \\
 PAT & 16.9 & 98.1 &  121.8\\
 NAT & 13.9 & 85.7 &  131.2\\
 EAT & 12.5 & 81.8 &  108.3\\
 \textcolor{black}{Free-4} & \textcolor{black}{11.1} & \textcolor{black}{92.5} & \textcolor{black}{115.6}\\
 \textbf{ANP} & 8.3 & \textbf{75.7}  & \textbf{79.2} \\
\bottomrule
\end{tabular}}
\end{sc}
\end{small}
\end{center}
%\vspace{-0.25in}
\end{table}

\subsection{Model Structure Robustness Evaluation}
In this section, we evaluate the structural robustness of deep models using our proposed metrics: \emph{\textcolor{black}{Empirical} Boundary Distance} and \emph{$\varepsilon$-Empirical Noise Insensitivity}.

\subsubsection{\textcolor{black}{Empirical} Boundary Distance}

\begin{figure}[!htb]
\centering
%\vspace{-0.15in}
%\hspace{-0.2in}
\subfigure[]{
\includegraphics[width=0.46\linewidth]{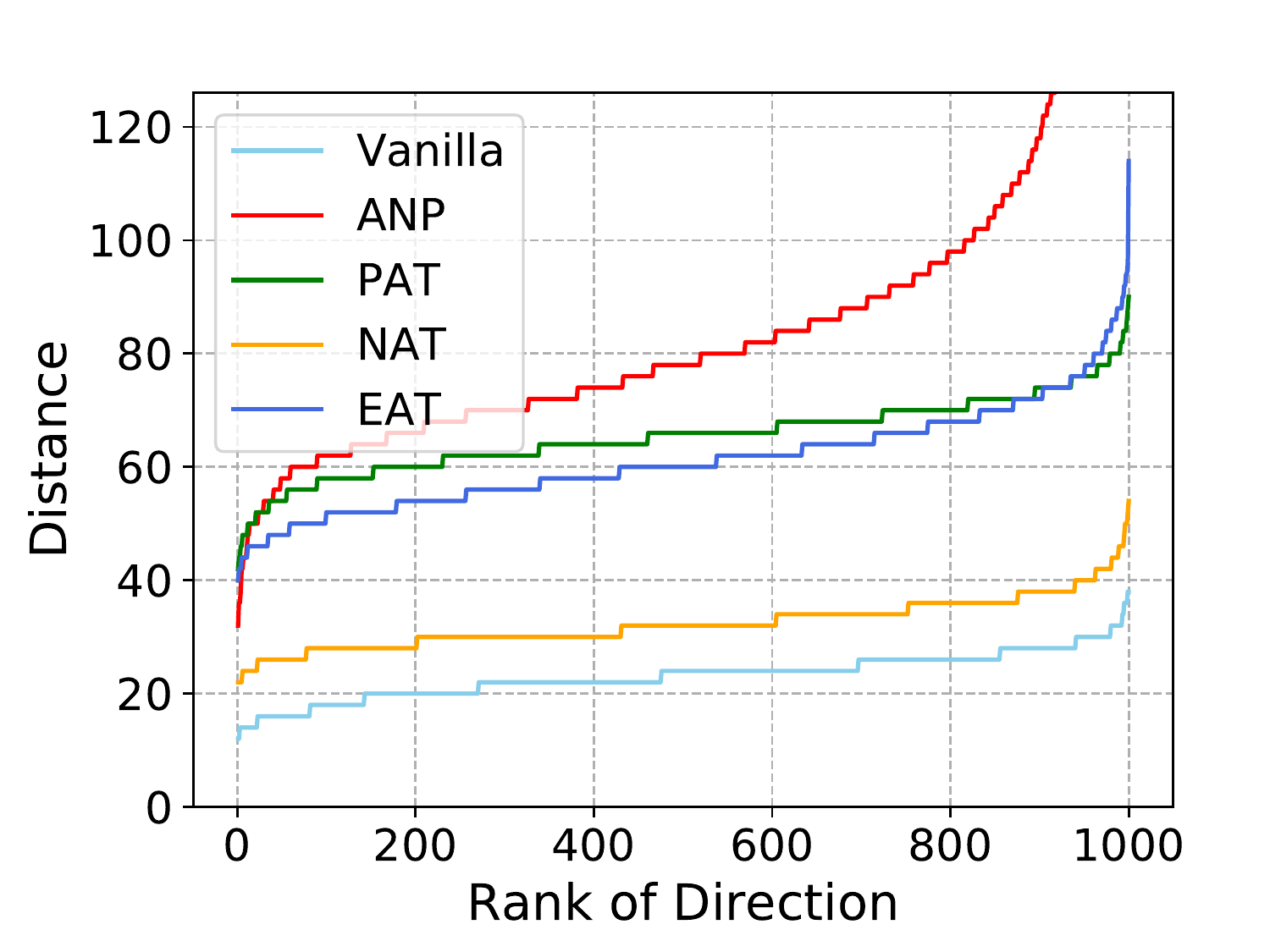}
}
%\hspace{-0.1in}
\subfigure[]{
\includegraphics[width=0.46\linewidth]{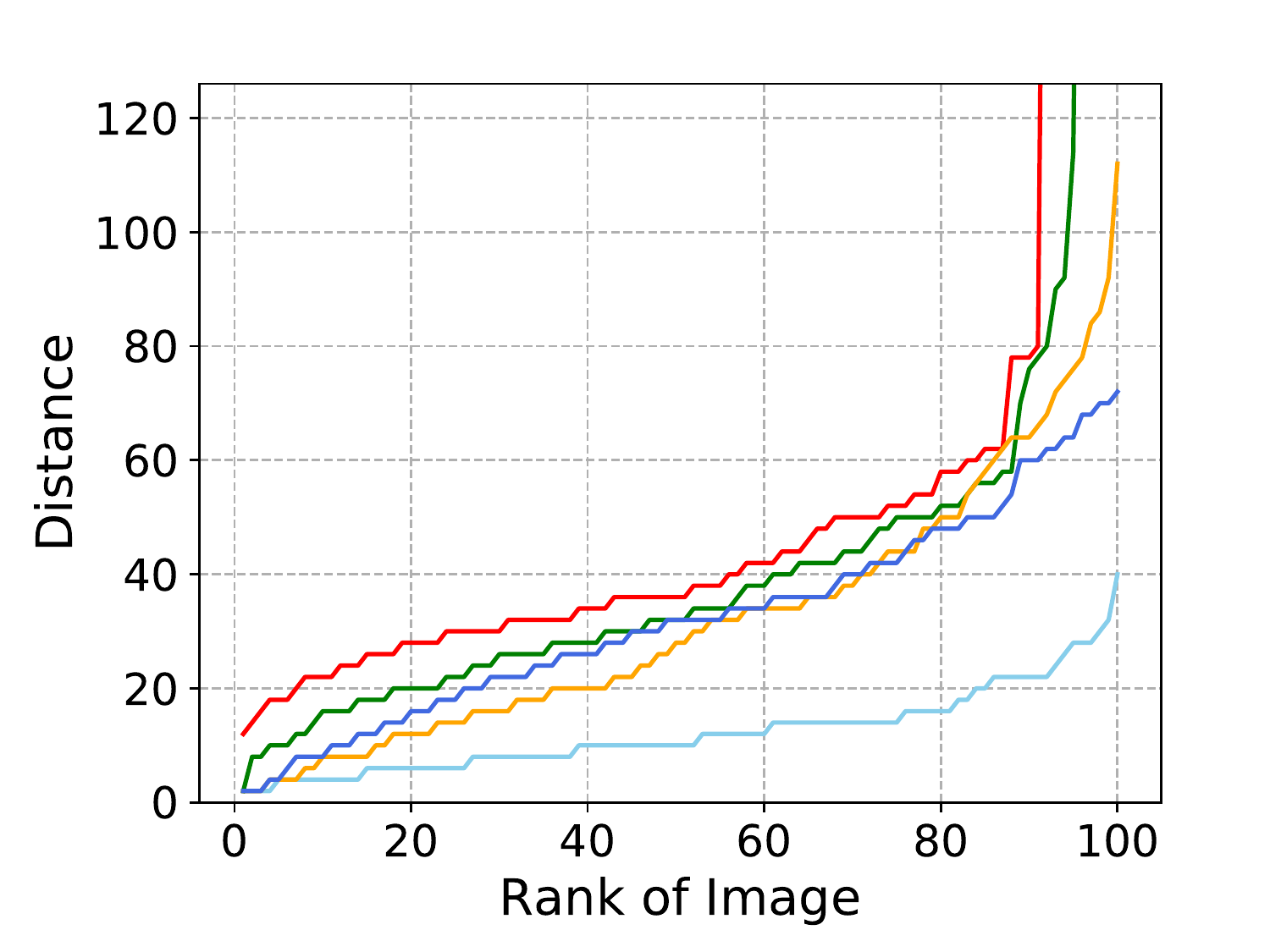}
}
%\vspace{0.0in}
\caption{\textcolor{black}{Empirical} Boundary Distance values are computed among five different VGG-16 models to change the predicted label: (a) the average distance moved in each orthogonal direction, and (b) the minimum distance (i.e., \textcolor{black}{Empirical} Boundary Distance) moved for 100 different images.}
\label{fig:wcdb}
%\vspace{-0.1in}
\end{figure}

We choose 1,000 randomly selected orthogonal directions, and compute the minimum distance along each direction for a specific image to change the predicted label. As can be seen from Figure \ref{fig:wcdb} (a), models trained by ANP have the largest distance. Moreover, Figure \ref{fig:wcdb} (b) provides the minimum distances moved for each of 100 randomly picked images in order to change their labels. It is easy to see that the distance curve of ANP is almost the highest, with a large leading gap at the beginning. Table \ref{tab:tab7} further reports $W_{f}$ figures for different methods. The results consistently prove that ANP supplies deep models with strong discriminating power by the largest margins, with the result that these models are the most robust.

\subsubsection{$\varepsilon$-Empirical Noise Insensitivity}

\begin{figure}[!htb]
\centering
%\vspace{-0.15in}
%\hspace{-0.2in}
\subfigure[adversarial examples]{
\includegraphics[width=0.46\linewidth]{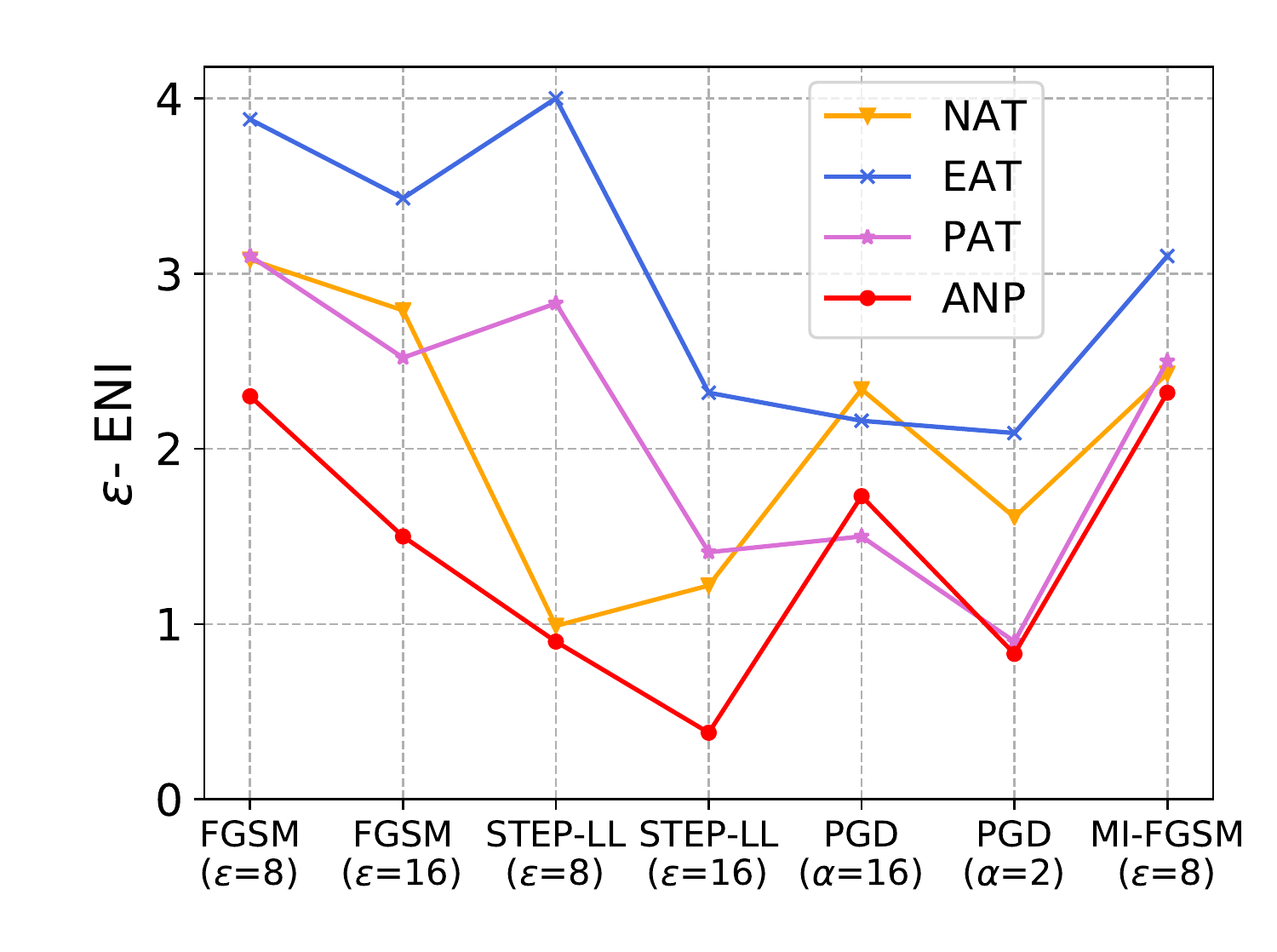}
}
%\hspace{-0.1in}
\subfigure[corruptions]{
\includegraphics[width=0.46\linewidth]{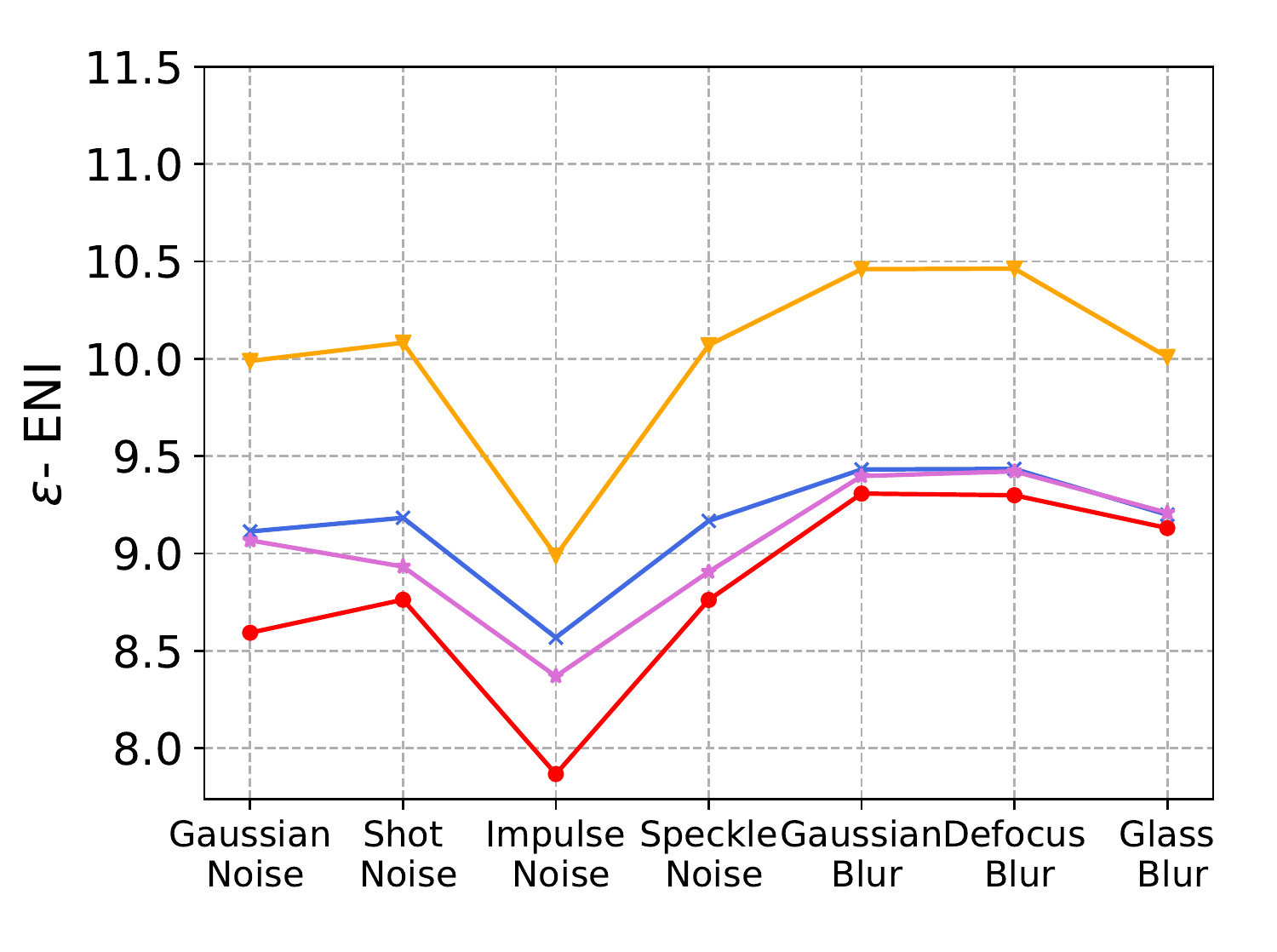}
}
\subfigure[FGSM]{
\includegraphics[width=0.46\linewidth]{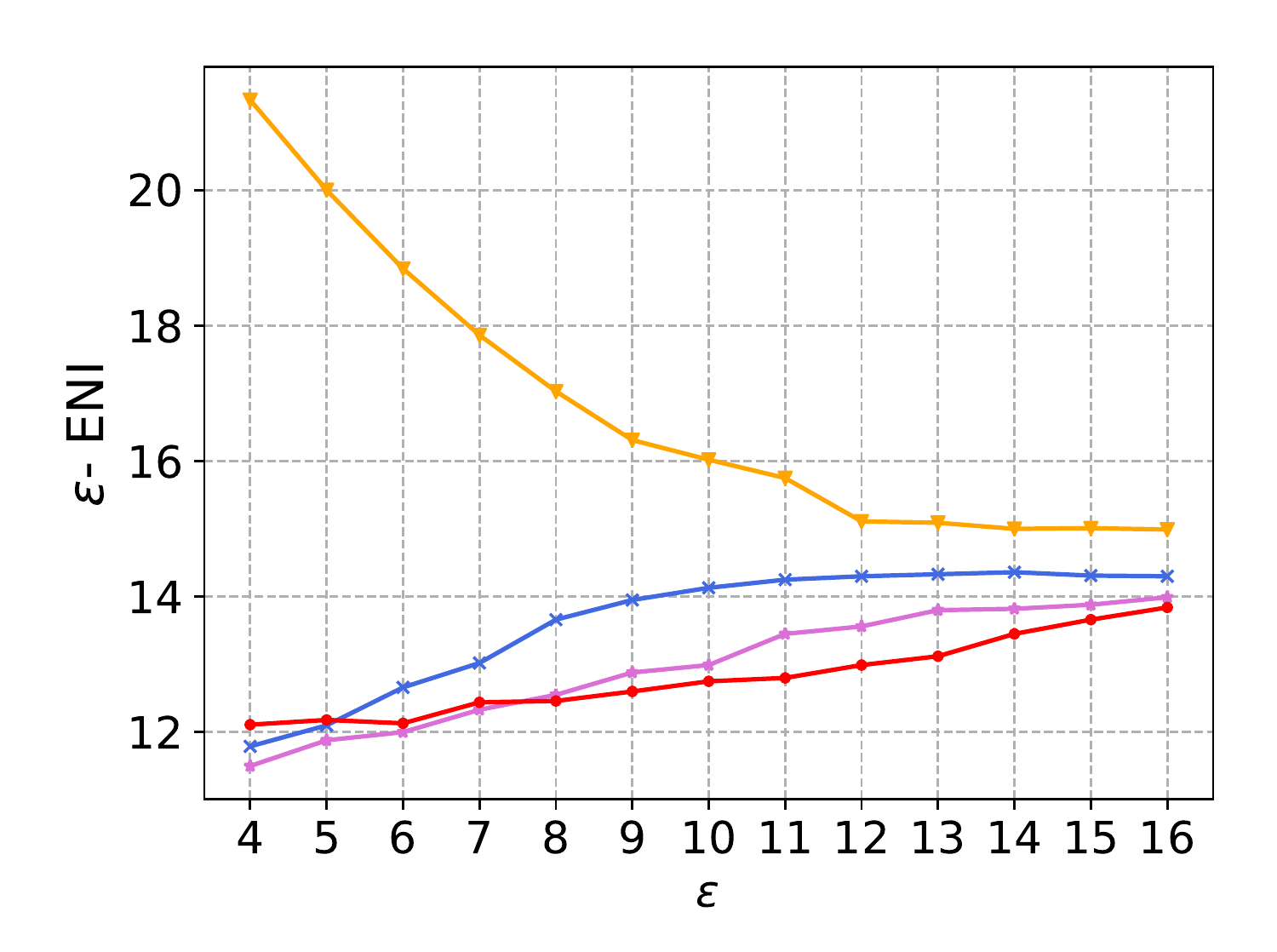}
}
%\hspace{-0.1in}
\subfigure[Step-LL]{
\includegraphics[width=0.46\linewidth]{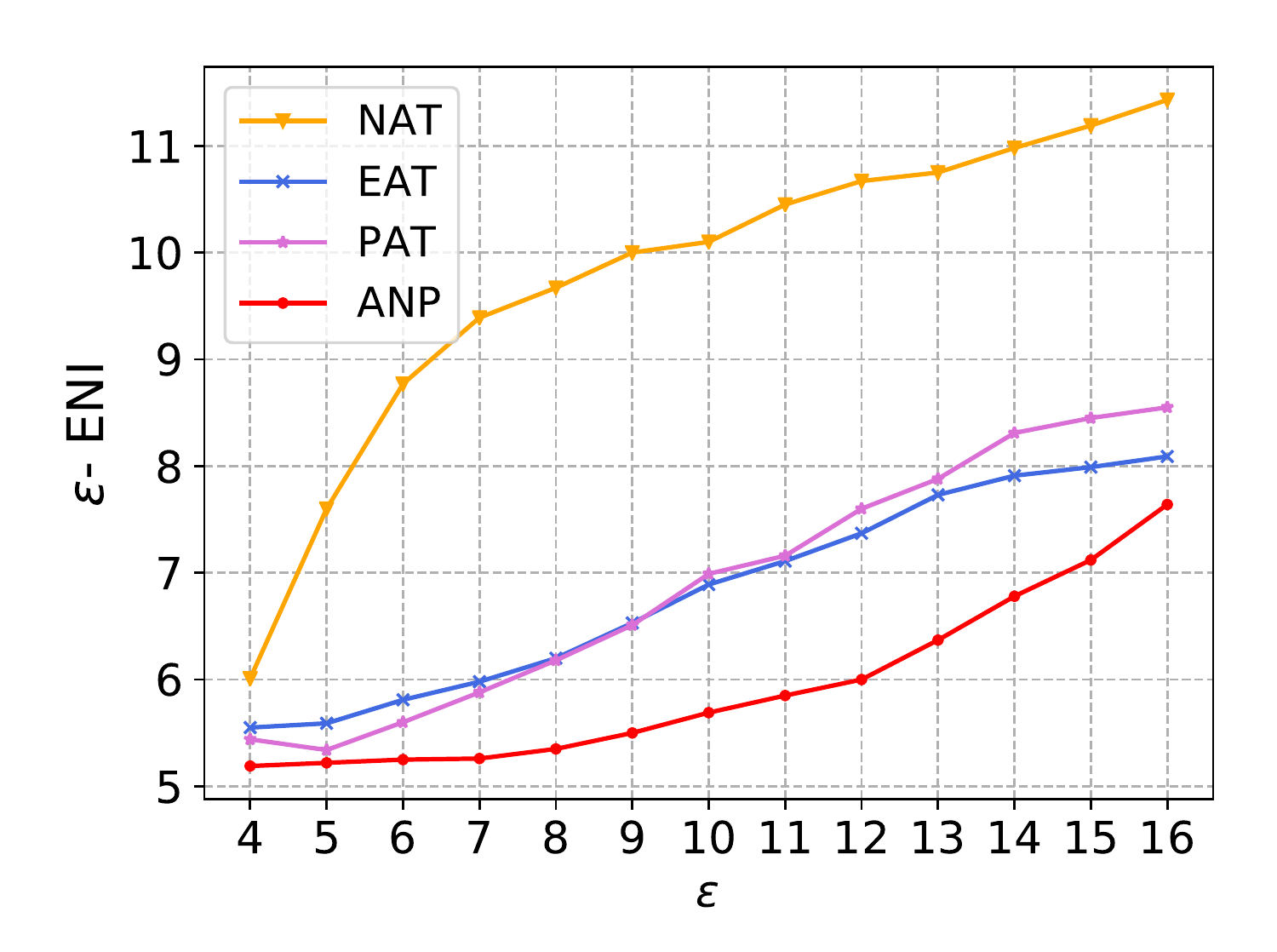}
}
\subfigure[Gaussian noise]{
\includegraphics[width=0.46\linewidth]{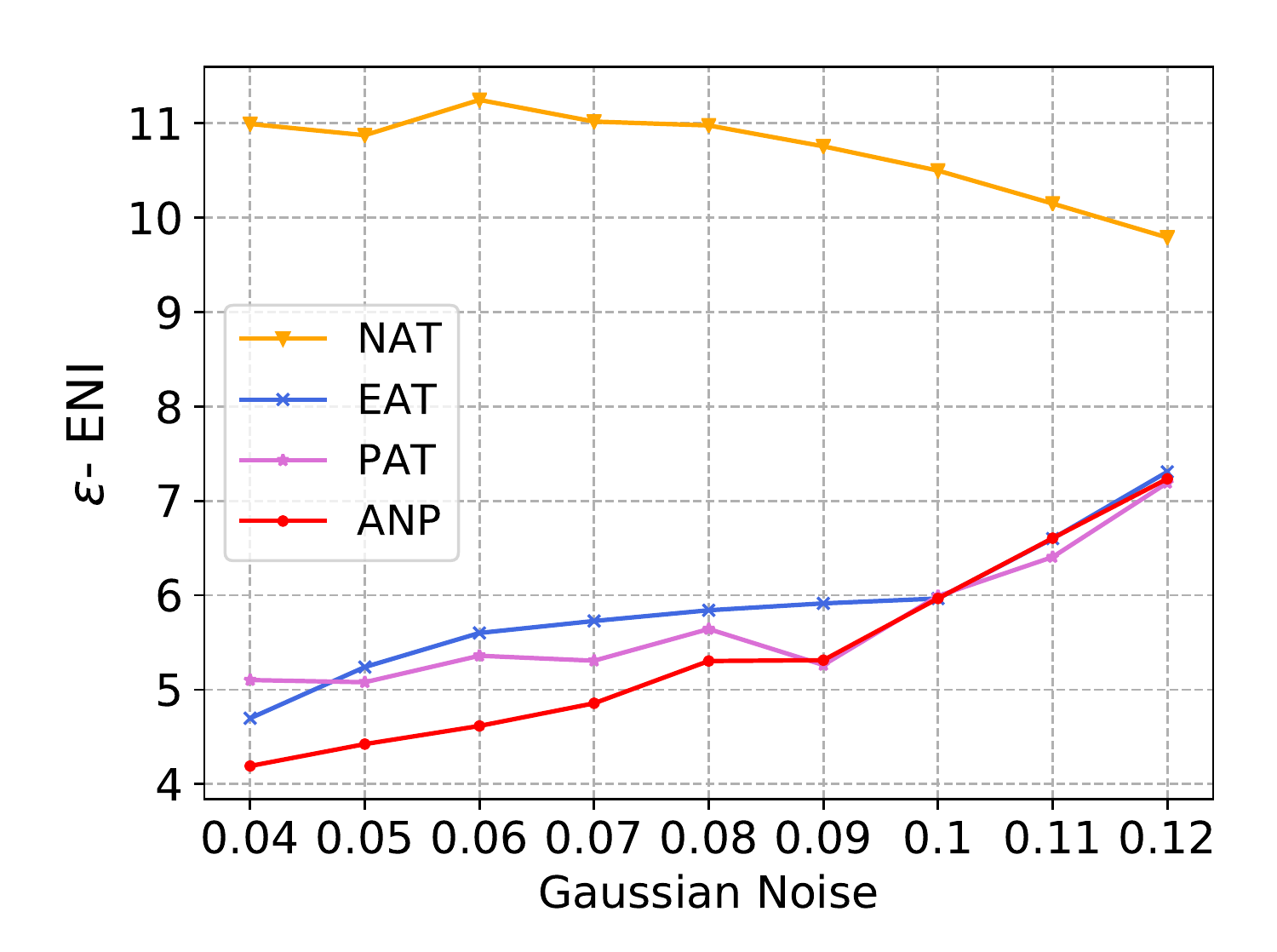}
}
%\hspace{-0.1in}
\subfigure[Shot noise]{
\includegraphics[width=0.46\linewidth]{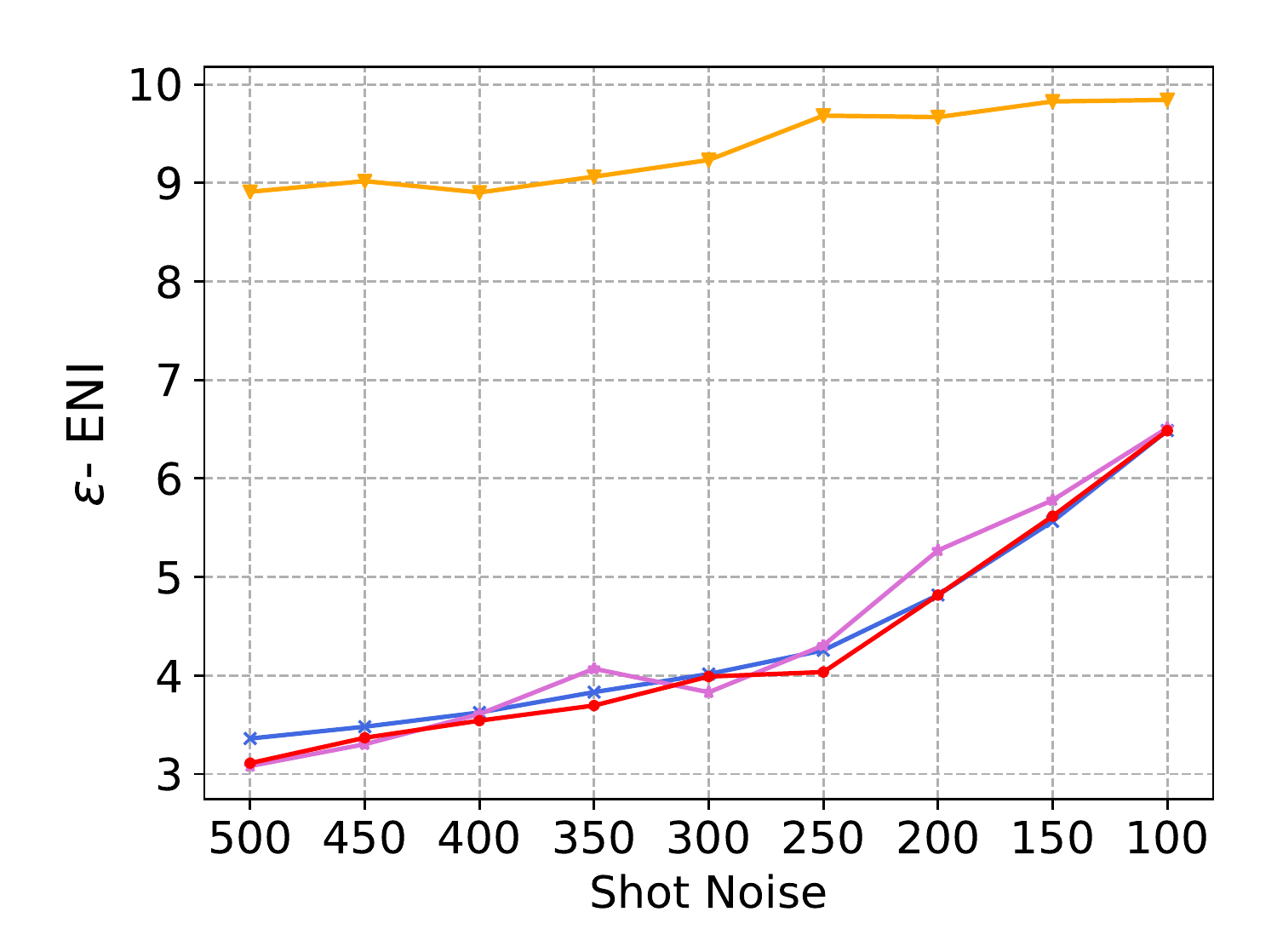}
}
\caption{$\varepsilon$-Empirical Noise Insensitivity is calculated under adversarial examples and corruptions.}
\label{fig:eni}

\end{figure}

In this section, several different methods (including FGSM, PGD, Gaussian noise, etc.) are employed to generate \textcolor{black}{polluted} examples from 100 clean images. For each clean image, 10 corresponding polluted examples are generated using every method within noise constraint $\varepsilon$. As shown in Figure \ref{fig:eni} (a) and (b), ANP obtains the smallest noise insensitivity in most cases, exhibiting strong robustness to both adversarial examples and corruption. More experimental results are shown in Figure \ref{fig:eni} (c) to (f). Specifically, Figure \ref{fig:eni} (c) and (d) illustrate the results on adversarial examples generated using FGSM and Step-LL; meanwhile, subfigures (e) and (f) show the results on Gaussian noise and Shot noise.

\begin{table}[!htb]
\vspace{-0in}
\caption{\textcolor{black}{Empirical} Boundary Distance among five VGG-16 models trained with different training methods. }
\label{tab:tab7}
\begin{center}
\begin{small}
\begin{sc}
\setlength{\tabcolsep}{1.1mm}{
\begin{tabular}{cccccc}
\toprule
Method & Vanilla & PAT & NAT & EAT & ANP \\
\midrule
$W_{f}$ & 29.46 & 42.23 & 34.22 & 35.21 & \textbf{47.36} \\
\bottomrule
\end{tabular}}
\end{sc}
\end{small}
\end{center}
%\vspace{-0.2in}
\end{table}


\subsection{Combination with Other Adversarial Defense Strategies}

Currently, most effective adversarial defense strategies involve adversarial training, in which adversarial examples are added during training. \cite{goodfellow6572explaining} first proposed adversarial training with an adversarial objective function based on the use of fast gradient sign method as an effective regularizer. Based on that, with the aim of universally robust networks, \cite{madry2017towards} trained models by feeding adversarially perturbed inputs with PGD into the loss term. \cite{NIPS2018_7324} proposed to improve adversarial robustness by integrating an adversarial perturbation-based regularizer into the classification objective. Recently, \cite{zhang2019theoretically} decomposed the prediction error for adversarial examples (robust error) into the sum of the natural (classification) error and boundary error, and consequently proposed the TRADES defense method that combines an accuracy loss and a regularization term for robustness.

Clearly, our proposed ANP framework takes advantage of hidden layers within a network and is orthogonal to the above mentioned adversarial training methods. \textcolor{black}{It is therefore intuitive for us to combine our ANP with these elaborately designed objective functions, as this could further improve model robustness by fully exploiting the potential of hidden layers. Accordingly, we further train models by combining ANP with TRADES (No.1 in \emph{NeurIPS 2018} adversarial defense competition) and PAT (the most commonly used adversarial training method).}

\textcolor{black}{As shown in Table \ref{tab:combination}, when ANP is combined with TRADES (ANP+TRADES) and PAT (ANP+PAT), this approach outperforms its counterparts (i.e., TRADES and PAT). More specifically, ANP+TRADES achieves the most robust performance and outperforms the compared methods. During training, we simply combine PAT or TRADES into our ANP framework in terms of the $\ell_{\infty}$ norm.} These rates enable us to draw an insightful conclusion that hidden layers could be considered and introduced along with other objective functions and regularizer terms to encourage stronger adversarial robustness in the future.

\subsection{What Did Hidden Layers Do during Noise Propagation?}
In this section, we aim to uncover the behavior and effect of hidden layers during noise propagation from the perspectives of hidden representation insensitivity and human vision alignment.

%\begin{figure}[!htb]
%\centering
%\includegraphics[width=0.9\linewidth]{res/layergroup_res18.pdf}
%\caption{Top-m layer group noise study with ResNet-18.}
%\label{fig:topm}
%\end{figure}

\begin{figure}[!htb]
\centering
%\vspace{-0.2in}
%\hspace{-0.15in}
\subfigure[]{
\includegraphics[width=0.46\linewidth]{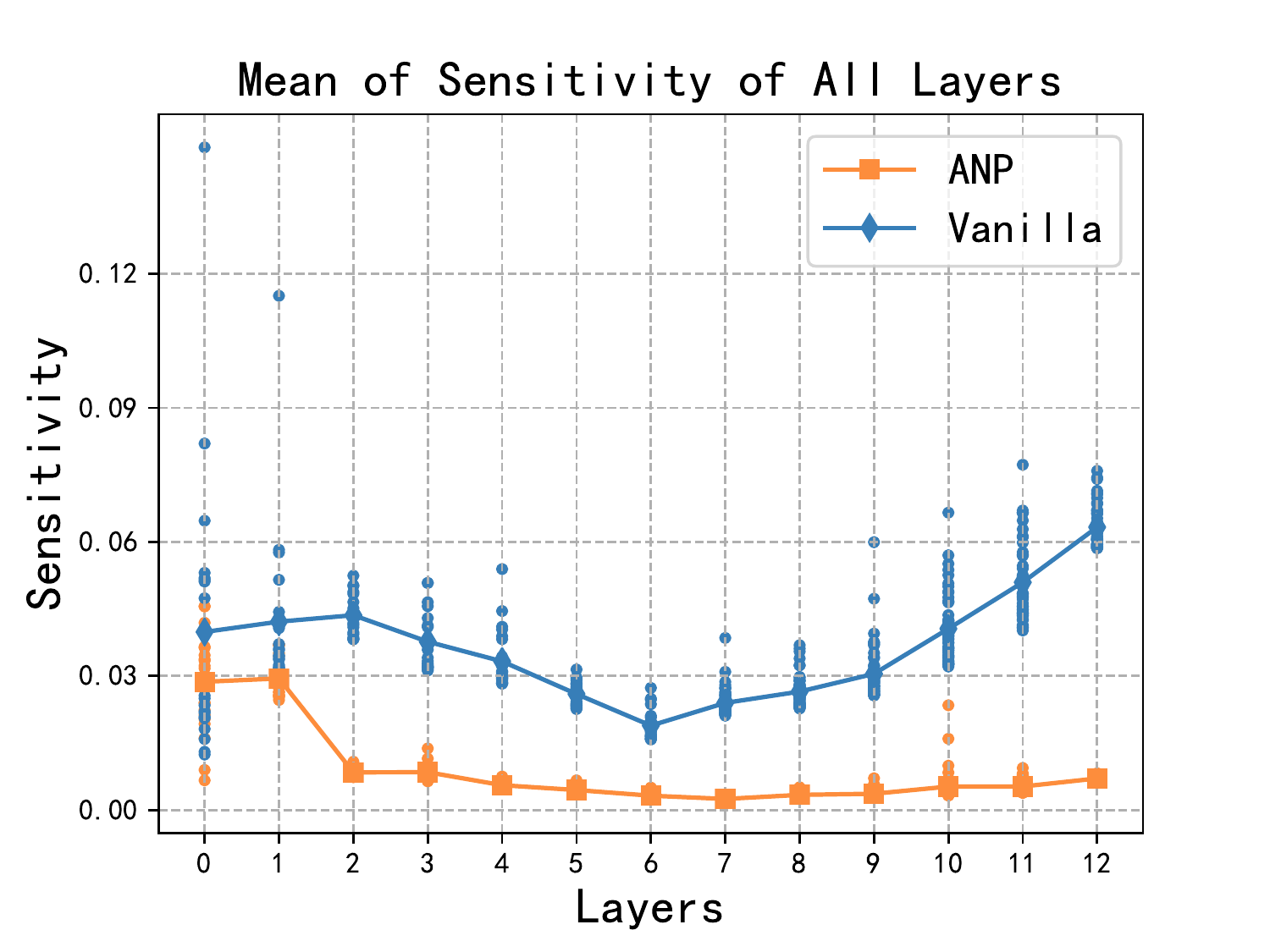}
}
%\hspace{-0.25in}
\subfigure[]{
\includegraphics[width=0.46\linewidth]{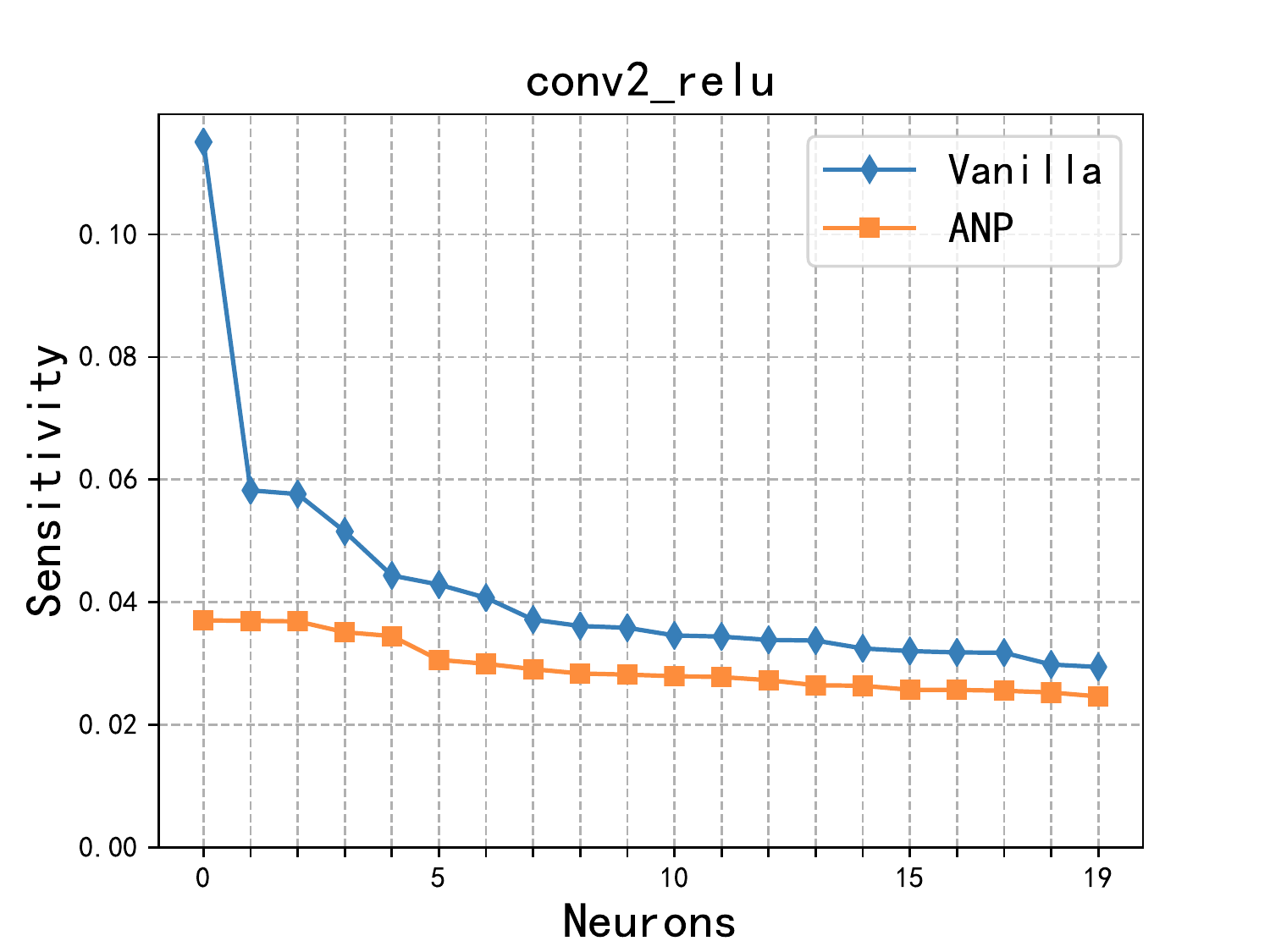}
}
\subfigure[]{
\includegraphics[width=0.46\linewidth]{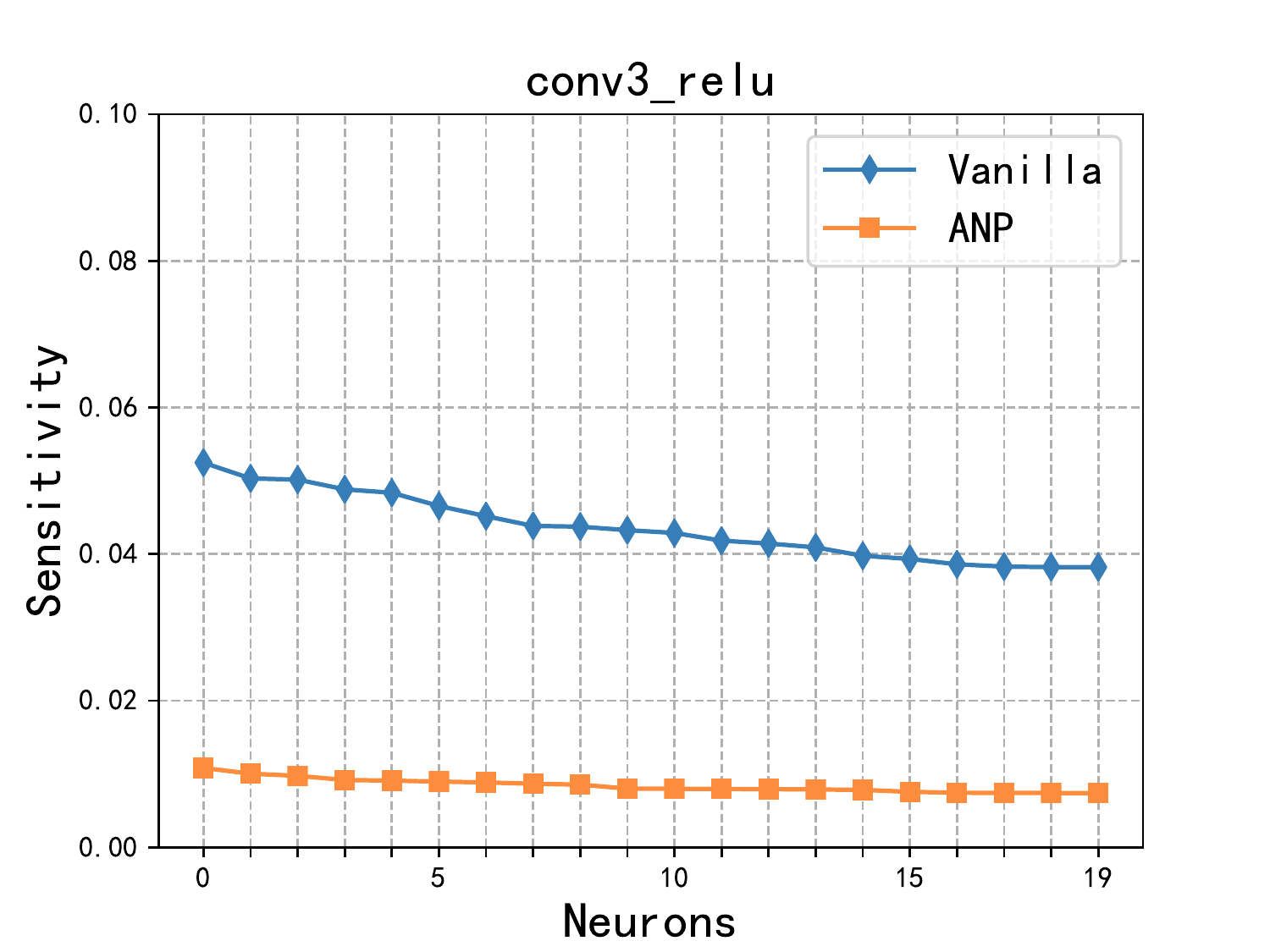}
}
%\hspace{-0.1in}
\subfigure[]{
\includegraphics[width=0.46\linewidth]{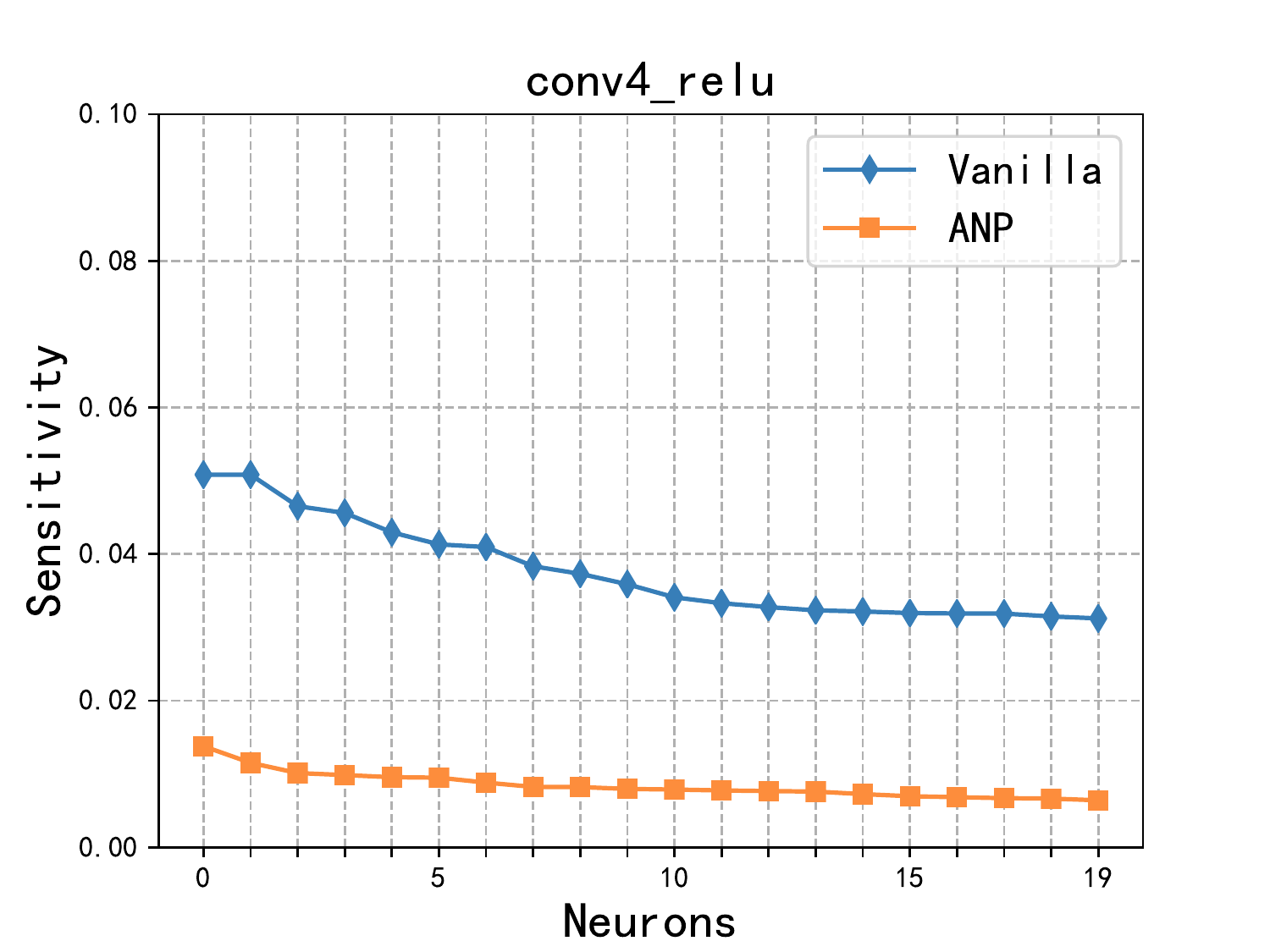}
}
\subfigure[]{
\includegraphics[width=0.46\linewidth]{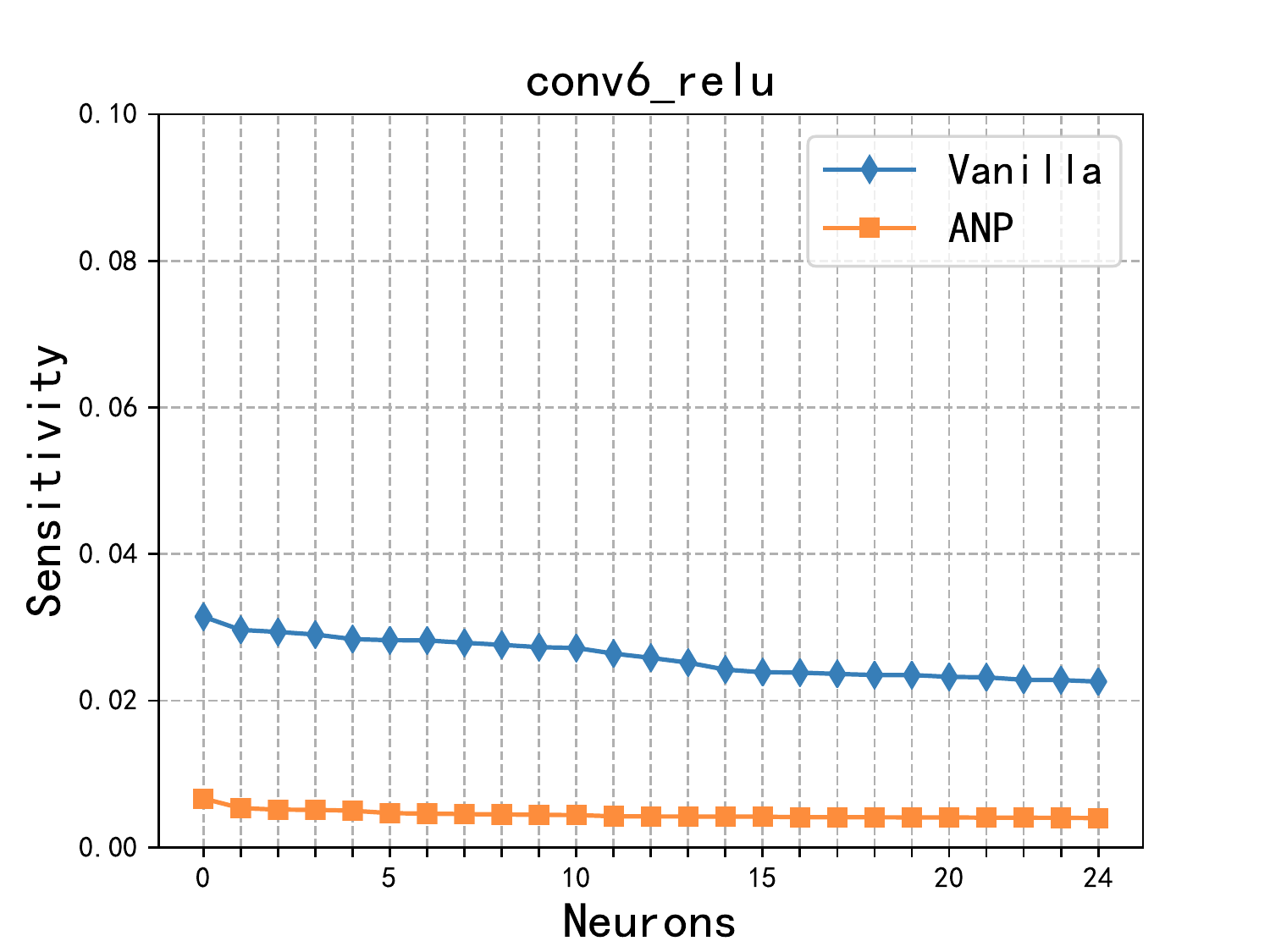}
}
%\hspace{-0.1in}
\subfigure[]{
\includegraphics[width=0.46\linewidth]{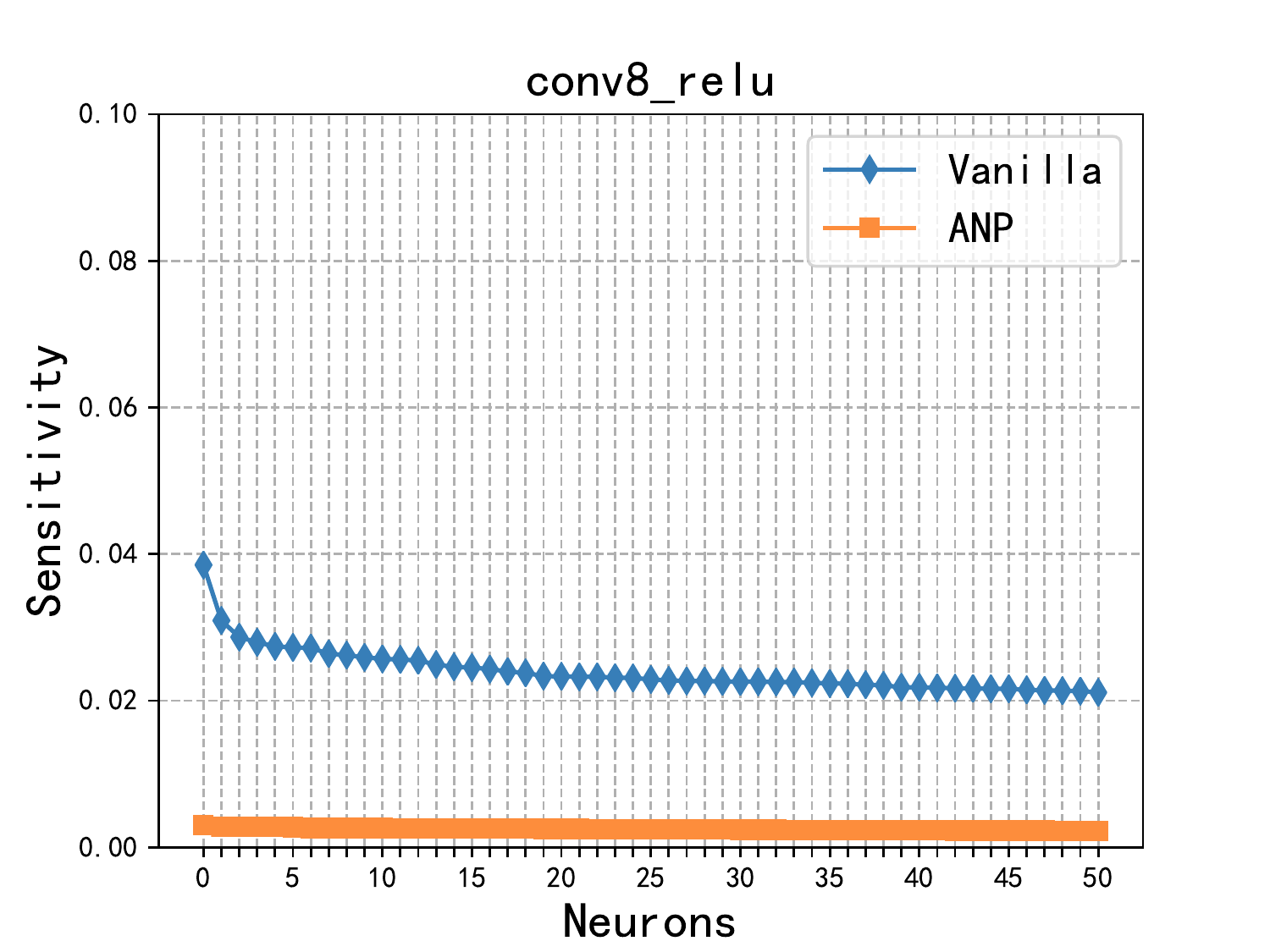}
}

\subfigure[]{
\includegraphics[width=0.46\linewidth]{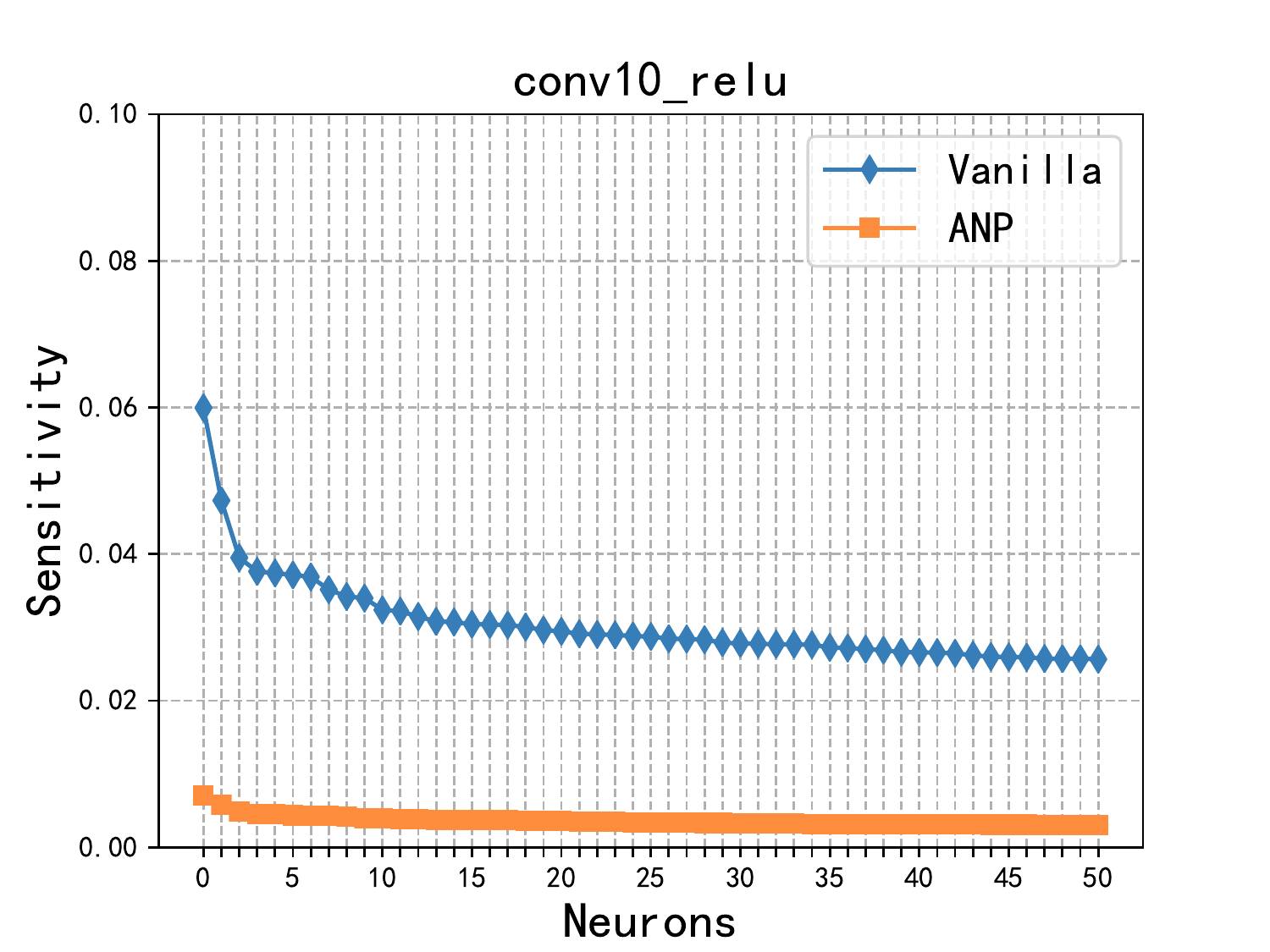}
}
%\hspace{-0.1in}
\subfigure[]{
\includegraphics[width=0.46\linewidth]{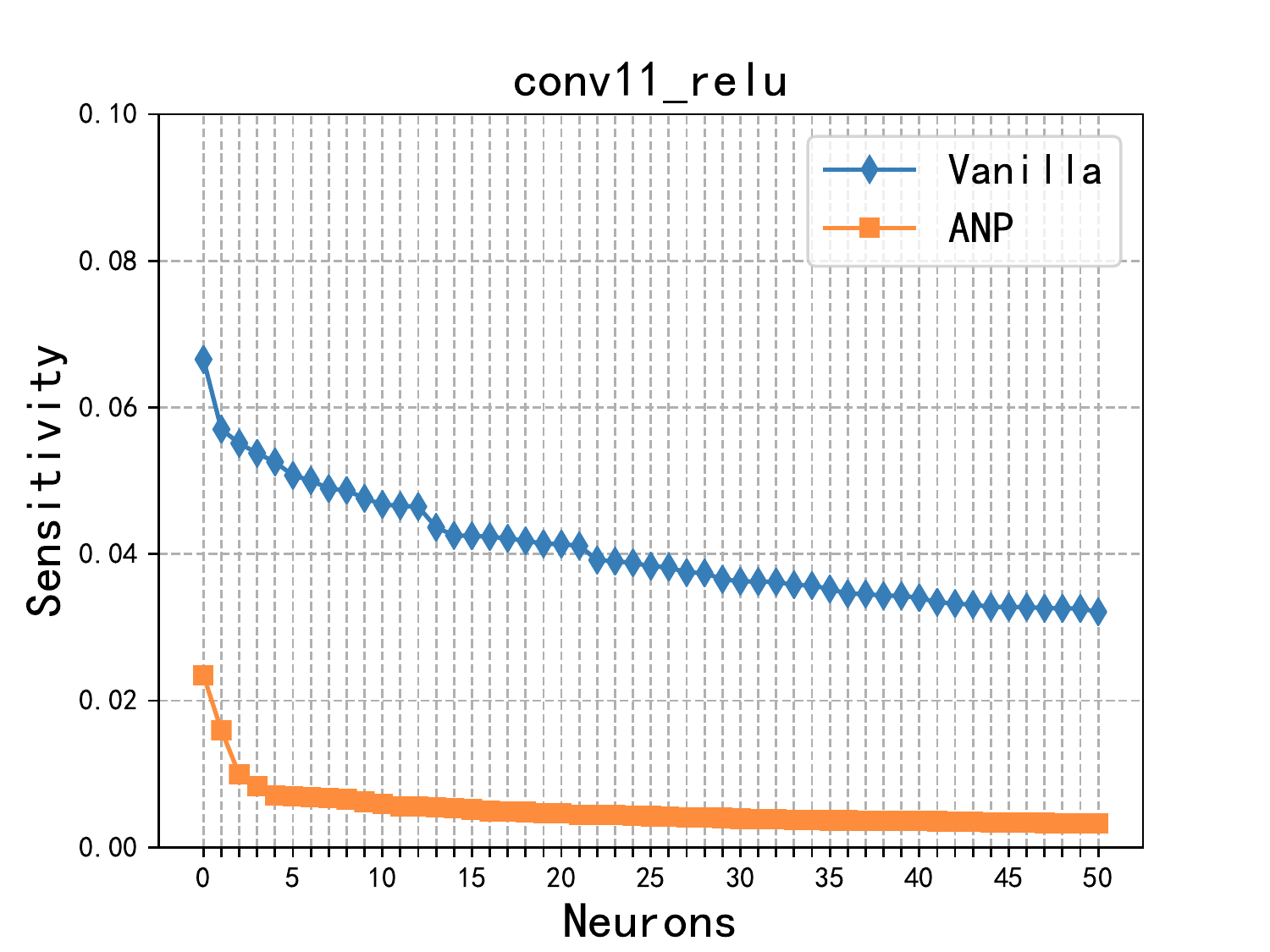}
}
\subfigure[]{
\includegraphics[width=0.46\linewidth]{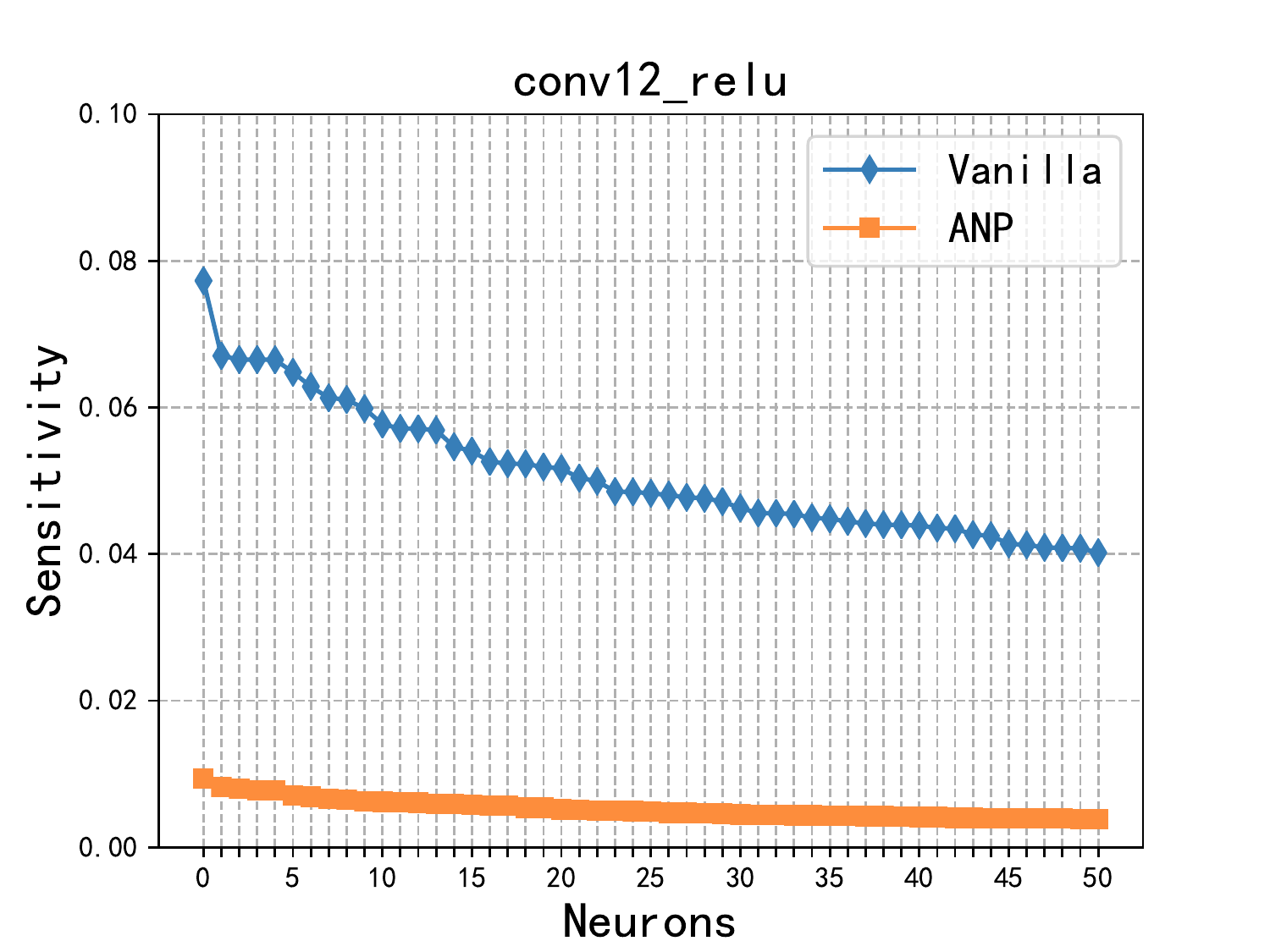}
}
%\hspace{-0.1in}
\subfigure[]{
\includegraphics[width=0.46\linewidth]{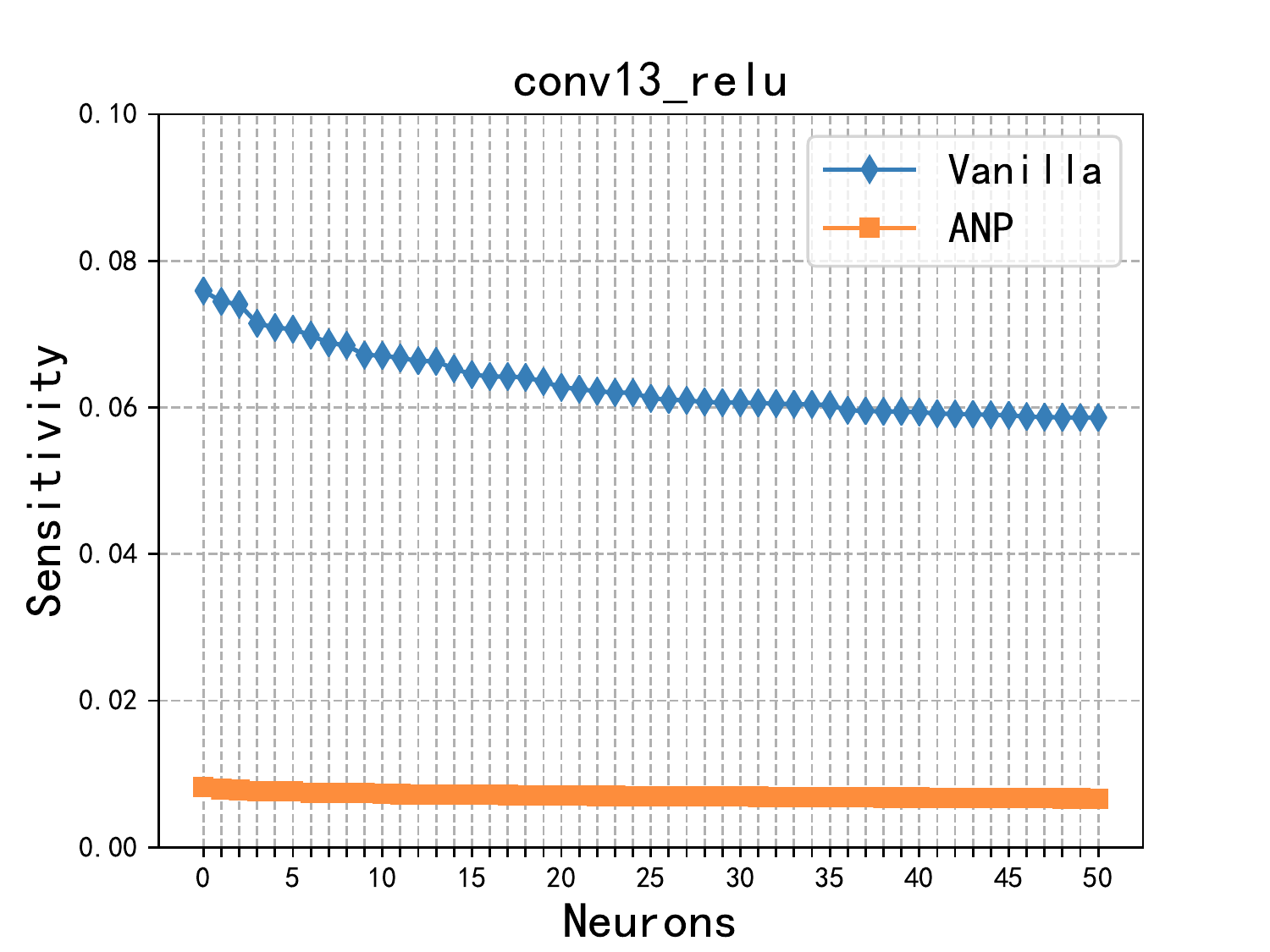}
}

%\hspace{-0.15in}
\caption{Hidden representation insensitivity on different layers. Subfigure (a) to (h) represent the mean for all layers and hidden representation insensitivity in layer \emph{conv2\_relu}, \emph{conv3\_relu}, \emph{conv4\_relu}, \emph{conv6\_relu}, \emph{conv8\_relu}, \emph{conv10\_relu}, \emph{conv11\_relu}, \emph{conv12\_relu} and \emph{conv13\_relu}, respectively.}
\label{fig:sensitivity}
%\vspace{-0.15in}
\end{figure}

Considered from a high-level perspective, robustness to \textcolor{black}{noise} can be viewed as a global insensitivity property that a model satisfies \cite{tsipras2018robustness}. A model that achieves small loss for \textcolor{black}{noise} in a dataset is necessarily one that learns representations that are insensitive to such \textcolor{black}{noise}. By injecting adversarial \textcolor{black}{noise} into hidden layers, ANP can be viewed as a method for embedding certain insensitivity into each hidden representation for models. We therefore try to explain the hidden layer behaviors from the perspective of hidden representation insensitivity. We measure the hidden representation insensitivity based on the degree of neuron activation value change within pairs of samples ($x$, $x'$), in which the distance between each pair is constrained with $\varepsilon$. Intuitively, when fed with benign and polluted examples, the more insensitively the neurons behave, the more robust the models will be. As shown in Figure \ref{fig:sensitivity}, neurons in each layer behave more insensitively to $\varepsilon$-noise (PGD attack adversarial examples and corruption) after being trained with ANP.

\begin{table}[!thb]
%\vspace{-0.1in}

\caption{White-box attack defense on CIFAR-10 with VGG-16.}
\label{tab:combination}
\begin{center}
\begin{small}
\begin{sc}
\setlength{\tabcolsep}{2.5mm}{
%\scriptsize
\begin{tabular}{ccccc}
\toprule
VGG-16 & \textcolor{black}{Clean} & FGSM & PGD & \textcolor{black}{BBAttack\cite{Brendel2019accurate}}\\
\cline{3-5}
 & &\scriptsize $\varepsilon$=8 & {\scriptsize $\varepsilon$=8} & \textcolor{black}{{\scriptsize $\varepsilon$=4}} \\
\midrule
 Vanilla & \textcolor{black}{\textbf{92.1\%}} & 1.4\% &  0.0\% & \textcolor{black}{2.3\%}\\
 TRADES & \textcolor{black}{83.2\%} & 47.9\% &  47.6\% & \textcolor{black}{64.7\%} \\
  \textcolor{black}{PAT} & \textcolor{black}{83.1\%} & \textcolor{black}{45.3\%} &  \textcolor{black}{41.4\%} & \textcolor{black}{61.2\%}\\
 \textbf{ANP}+\textbf{TRADES} & \textcolor{black}{82.1\%} & \textbf{48.5\%} &  \textbf{49.8\%} & \textbf{65.8\%}\\
  \textcolor{black}{\textbf{ANP}+\textbf{PAT}} & \textcolor{black}{82.0\%} & \textcolor{black}{46.9\%} &  \textcolor{black}{44.3\%} & \textcolor{black}{62.9\%}\\
\bottomrule
\end{tabular}}
\end{sc}
\end{small}
\end{center}
%\vspace{-0.25in}
\end{table}

\begin{figure}[!htb]
\centering
\subfigure[]{
\begin{minipage}[b]{0.16\linewidth}
\includegraphics[width=1\linewidth]{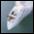}\vspace{0.02in}
\includegraphics[width=1\linewidth]{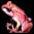}\vspace{0.02in}
\includegraphics[width=1\linewidth]{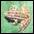}\vspace{0.02in}
\includegraphics[width=1\linewidth]{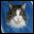}\vspace{0.02in}
\includegraphics[width=1\linewidth]{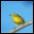}\vspace{0.02in}
\includegraphics[width=1\linewidth]{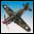}\vspace{0.02in}
\includegraphics[width=1\linewidth]{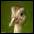}\vspace{0.02in}
\end{minipage}
}
%\hspace{-0.05in}
\subfigure[]{
\begin{minipage}[b]{0.16\linewidth}
\includegraphics[width=1\linewidth]{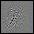}\vspace{0.02in}
\includegraphics[width=1\linewidth]{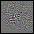}\vspace{0.02in}
\includegraphics[width=1\linewidth]{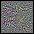}\vspace{0.02in}
\includegraphics[width=1\linewidth]{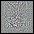}\vspace{0.02in}
\includegraphics[width=1\linewidth]{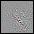}\vspace{0.02in}
\includegraphics[width=1\linewidth]{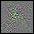}\vspace{0.02in}
\includegraphics[width=1\linewidth]{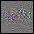}\vspace{0.02in}
\end{minipage}
}
%\hspace{-0.05in}
\subfigure[]{
\begin{minipage}[b]{0.16\linewidth}
\includegraphics[width=1\linewidth]{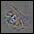}\vspace{0.02in}
\includegraphics[width=1\linewidth]{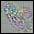}\vspace{0.02in}
\includegraphics[width=1\linewidth]{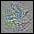}\vspace{0.02in}
\includegraphics[width=1\linewidth]{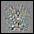}\vspace{0.02in}
\includegraphics[width=1\linewidth]{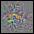}\vspace{0.02in}
\includegraphics[width=1\linewidth]{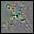}\vspace{0.02in}
\includegraphics[width=1\linewidth]{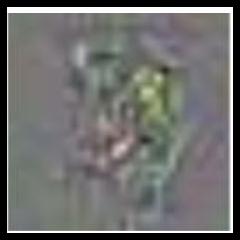}\vspace{0.02in}
\end{minipage}
}
%\hspace{-0.05in}
\subfigure[]{
\begin{minipage}[b]{0.16\linewidth}
\includegraphics[width=1\linewidth]{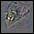}\vspace{0.02in}
\includegraphics[width=1\linewidth]{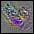}\vspace{0.02in}
\includegraphics[width=1\linewidth]{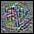}\vspace{0.02in}
\includegraphics[width=1\linewidth]{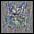}\vspace{0.02in}
\includegraphics[width=1\linewidth]{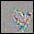}\vspace{0.02in}
\includegraphics[width=1\linewidth]{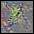}\vspace{0.02in}
\includegraphics[width=1\linewidth]{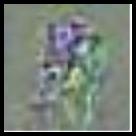}\vspace{0.02in}
\end{minipage}
}
%\vspace{-0.15in}
\caption{Visualization of the loss gradient w.r.t. input pixels on CIFAR-10. Subfigure (a) denotes the input image, (b) to (d) represent gradients gained from Vanilla model, ANP on top-all layers and ANP on top-4 layers, respectively. No preprocessing was applied to the gradients (other than scaling and clipping for visualization).}
\label{fig:gradient}
%\vspace{-0.15in}
\end{figure}

Finally, we explain the model robustness from the perspective of alignment with human visual perception, such that a stronger model gains a more semantically meaningful gradient. As shown in Figure \ref{fig:gradient} (a) to (d), gradients for ANP-trained networks on top-4 layers (d) align well with the perceptually relevant features (such as edges) of the input images. By contrast, these gradients have no coherent patterns and appear very noisy to humans for vanilla networks (b), and are less semantically meaningful for the ANP-trained model on all layers (c).

%\subsection{Adding Noises to Different Parts of the Network: A Case Study on ResNet-18.}
%
%
%\begin{table}[!thb]
%%\vspace{-0.1in}
%\caption{White-box attack defense on CIFAR-10 with VGG-16.}
%\label{tab:resblock}
%\begin{center}
%\begin{small}
%\begin{sc}
%\setlength{\tabcolsep}{2.5mm}{
%\begin{tabular}{cccc}
%\toprule
%ResNet-16 & BPDA & PGD & Gaussian \\
%\cline{2-4}
% &\scriptsize $\varepsilon$=8 & {\scriptsize $\varepsilon$=8} &  \\
%\midrule
% Vanilla & 51.6\% & 27.1\% &  50.4\% \\
% TRADES & \textbf{54.0\%} & \textbf{29.0\%} &  \textbf{68.3\%}\\
%\bottomrule
%\end{tabular}}
%\end{sc}
%\end{small}
%\end{center}
%%\vspace{-0.25in}
%\end{table}
%
%In this section, we bring experiments for models trained by ANP with different depths, i.e., numbers of layers. We use ANP to train VGG-11, VGG-13, VGG-16 and VGG-19 with fixed hyper-parameters and test their performance on clean examples, adversarial examples and corruption, respectively. Experimental results in Figure \ref{fig:shallowdeep} show that the classification accuracy on clean and adversarial examples as well as negative mCE on corruption for 4 different models are nearly the same which indicates that models trained with ANP shows barely superiority between shallow and deep architectures against generalized noises.

% needed in second column of first page if using \IEEEpubid
%\IEEEpubidadjcol

%\subsubsection{Subsubsection Heading Here}
%Subsubsection text here.

%\section{Conclusion}
%The conclusion goes here.
\section{Conclusion}
In order to improve model robustness against \textcolor{black}{noise}, this paper proposes a novel training strategy, named \emph{Adversarial Noise Propagation} (ANP), which injects diversified \textcolor{black}{noise} into the hidden layers in a layer-wise manner. ANP can be efficiently implemented through the standard backward-forward process, meaning that it introduces no additional computations. \textcolor{black}{Our study of the behaviors of hidden layers yielded two significant conclusions: (1) empirical studies reveal that we only need to perturb shallow layers to train robust models; (2) theoretical proof demonstrates that the shallow layers have stronger negative influences than deep layers in terms of clean accuracy.} Extensive experiments on the visual classification task demonstrate that ANP can enable strong robustness for deep networks, and can therefore aid in obtaining very promising performance against various types of noise.

Currently, most successful adversarial defense strategies considered only the input layers and trained models with elaborately designed loss functions and regularization terms. Fortunately, our proposed ANP framework takes advantage of
hidden layers and is orthogonal to most of these adversarial training methods, which can be combined together to build stronger models. \textcolor{black}{According to our experimental results, we can obtain a stronger model when ANP is provided with other adversarial training methods.} Thus, further researchers could propose training methods that fully exploit the potential of hidden layers and can therefore promise stronger adversarial robustness.

Furthermore, our current strategy injects the same amount of adversarial \textcolor{black}{noise} into each layer and largely requires manual efforts (e.g., the magnitude of \textcolor{black}{noise} or specific layers). However, since every layer contributes to the model robustness to a different extent, it is preferable for us to devise a more adaptive algorithm that considers the heterogeneous behaviors of different layers. It would therefore fully exploit the efficacy of every hidden layer and facilitate the building of stronger models. Meanwhile, it is intriguing to see that adding \textcolor{black}{noise} to \textcolor{black}{shallow} layers improves model robustness, while \textcolor{black}{perturbing the deep} layers has the opposite effect. The reasons for this remain unclear; we will investigate these further in future work.

\bibliographystyle{IEEEtran}
% argument is your BibTeX string definitions and bibliography database(s)
\bibliography{main}
\end{document}